\documentclass[letterpaper, 10 pt, conference]{ieeeconf} 

\IEEEoverridecommandlockouts 
\usepackage[pdftex]{graphicx}
\usepackage{amsmath}
\usepackage{arydshln}
\usepackage{hhline}
\usepackage{eucal}
\usepackage{cite}
\usepackage{amssymb}
\usepackage{enumerate}
\usepackage{subfigure}
\usepackage{bm}
\usepackage{color}
\usepackage{multirow}
\usepackage{algorithm} 
\usepackage{algpseudocode} 
\usepackage{graphicx}
\usepackage{svg}
\usepackage{booktabs}

\makeatletter
\renewcommand{\ALG@beginalgorithmic}{\small}
\makeatother

\newtheorem{rem}{Remark}

\DeclareSymbolFontAlphabet{\mathcal}   {symbols}

\title{\LARGE \bf
Proprioceptive Sensor-Based Simultaneous Multi-Contact Point Localization and Force Identification for Robotic Arms
}

\author{Seo Wook Han$^{1}$ and Min Jun Kim$^{1}$
\thanks{This work was supported by the National Research Foundation of Korea (NRF) grant funded by the Korea government (MSIT) No. 2021R1C1C1005232, and No. 2021R1A4A3032834.
\textsuperscript{1}The authors are with Intelligent Robotic Systems Laboratory, Korea Advanced Institute of Science and Technology, Daejeon, Republic of Korea. {E-mail: {\tt\small tjdnr7117, minjun.kim@kaist.ac.kr}}}
}

\begin{document}

\maketitle
\thispagestyle{empty}
\pagestyle{empty}

\begin{abstract}
In this paper, we propose an algorithm that estimates contact point and force simultaneously. We consider a collaborative robot equipped with proprioceptive sensors, in particular, joint torque sensors (JTSs) and a base force/torque (F/T) sensor. The proposed method has the following advantages. First, fast computation is achieved by proper preprocessing of robot meshes. Second, multi-contact can be identified with the aid of the base F/T sensor, while this is challenging when the robot is equipped with only JTSs. The proposed method is a modification of the standard particle filter to cope with mesh preprocessing and with available sensor data. In simulation validation, for a 7 degree-of-freedom robot, the algorithm runs at 2200$\mathrm{Hz}$ with 99.96\% success rate for the single-contact case. In terms of the run-time, the proposed method was $\geq$3.5X faster compared to the existing methods. Dual and triple contacts are also reported in the manuscript.
\end{abstract} 

\section{Introduction}
\label{sec:introduction}

Collaborative robots, which are becoming more and more popular nowadays, are designed to share a workspace with humans. Consequently, safe physical interaction under a collision or a contact has become an important research topic.
In some studies, for example, collision-free path planning \cite{ebert2002safe,kuhn2007fast,kim2023tamp} and passivity-based controller \cite{kim2019passivity, kim2016passivity} are used to ensure safety. Some other studies aim at identifying occurrence of a collision as quick as possible for the sake of safety \cite{de2006collision,kim2015design,heo2019collision,birjandi2020observer, kim2021transferable, park2021collision}.

\begin{figure}[tb!]
    \centering
    \subfigure[]
	{\includegraphics[width=0.235\textwidth]{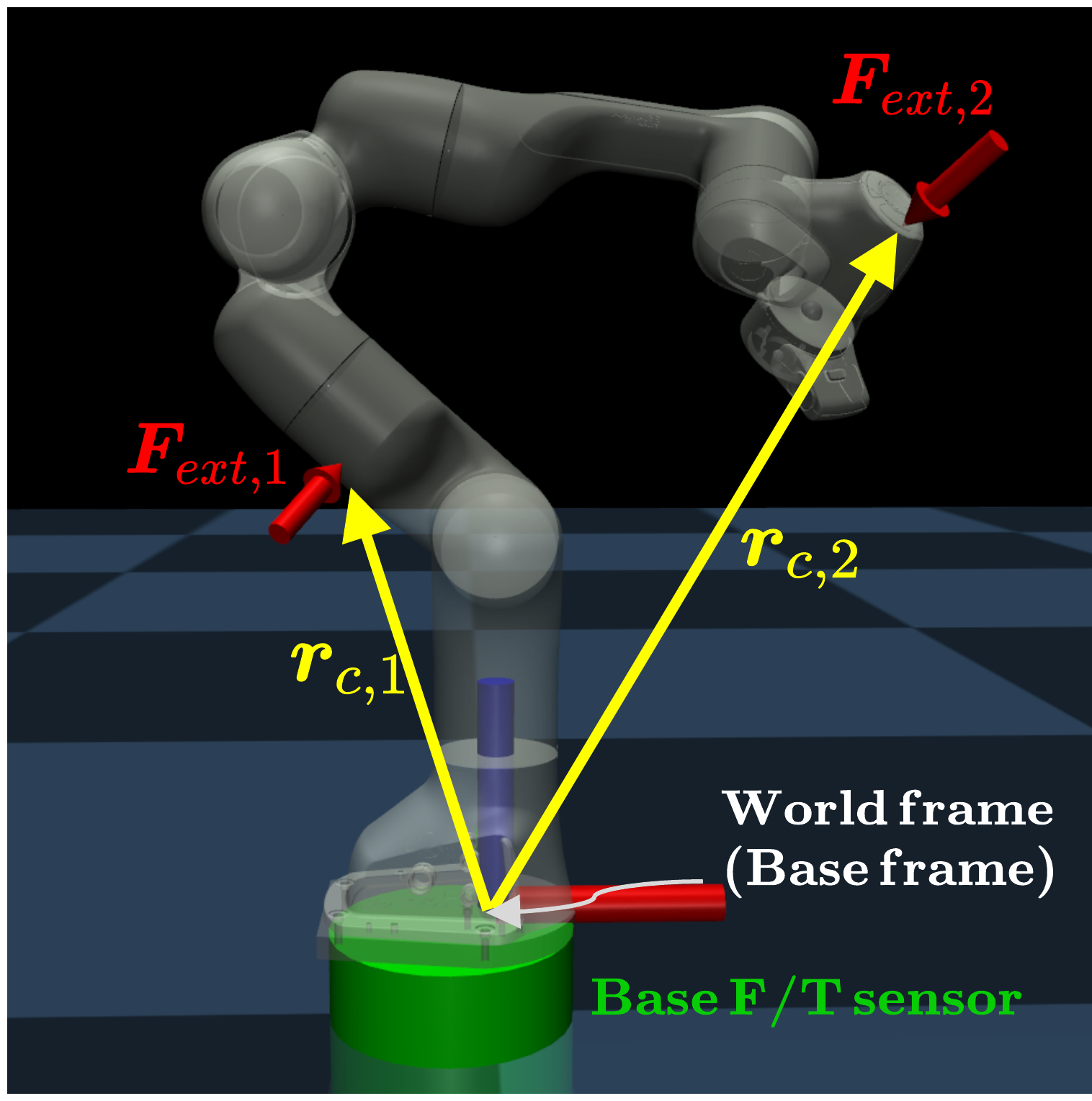}}
	\subfigure[]
	{\includegraphics[width=0.235\textwidth]{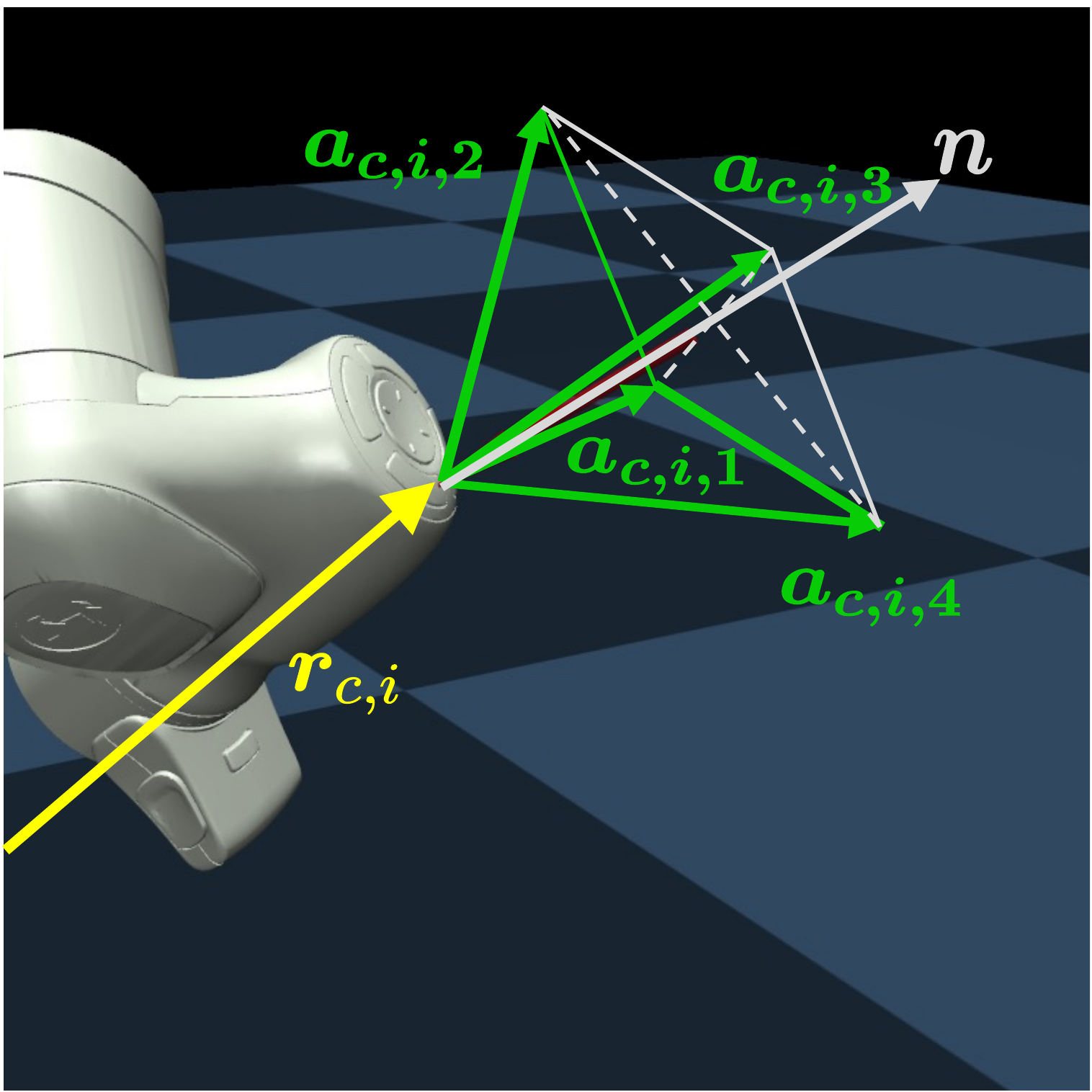}}
    \caption{
    (a) The goal of this paper is to estimating contact location ($\boldsymbol{r}_{c,i} \in \mathbb{R}^3$) and force ($\boldsymbol{F}_{ext,i} \in \mathbb{R}^3$) using proprioceptive sensors (JTS, base F/T sensor) (b) Point contact assumption is realized by friction cone formulation.
    }
    \vspace{-5mm}
    \label{fig:vector_convention}
\end{figure}

To further understand and treat the physical interaction properly, there have been attempts to estimate contact location and force simultaneously (e.g., estimating $\boldsymbol{r}_{c,i}$ and $\boldsymbol{F}_{ext,i}$ in Fig. \ref{fig:vector_convention}(a)). 
For this problem, a major branch of research is to develop tactile sensors that try to mimic sensory receptors of the human skin \cite{8695719, 8932392}. Although tactile sensors are a very promising technology, a more practical setup to date would be to use proprioceptive sensors, such as a joint torque sensor (JTS) or a 6-axis force/torque (F/T) sensor, which are already mature and available in the market.

Indeed, a number of solutions have been proposed using proprioceptive sensors \cite{zwiener2018contact,8059840,buondonno2016combining, iskandar2021collision, pang2021identifying, popov2021real}.
For example, \cite{zwiener2018contact} proposed a method based on the machine learning technique, and \cite{8059840} proposed a so-called line of force action method that computes a contact location geometrically. These methods, however, are known to have limited identifiable contact locations. 
\cite{zwiener2018contact} considers only a few discretized points on robot link, and the methods in \cite{8059840, buondonno2016combining, iskandar2021collision} may have no feasible solutions depending on the link geometry.

To estimate the contacts in a more general setup, particle filter (PF)-based approaches have been proposed in \cite{manuelli2016localizing,zwiener2019armcl,bimbo2019collision}. In this approach, particles are updated and resampled to find a pair of location and force that explains all sensors measurements, in the sense of a quadratic programming (QP). Moreover, within the framework of PF, multi-contact can be estimated when sensory information is rich enough.

However, a couple of challenges remain in the PF-based approaches. First, the run-time is typically slow, 
mainly due to the computation time in the motion model of PF; e.g., a motion model proposed in \cite{manuelli2016localizing} projects particles to the robot surface after updating them in 3D space.
The run-time reported in \cite{manuelli2016localizing} is about 10$\mathrm{Hz}$ for a 30 degrees-of-freedom (DOF) humanoid robot. \cite{zwiener2019armcl} also proposed a PF-based method, resulting in about 200$\mathrm{Hz}$ for a 7-DOF robot arm.
A method proposed in \cite{popov2021real}, which is in fact not a PF variant to be precise, showed that the run-time can be improved with a proper preprocessing of robot meshes, demonstrating $\geq$600$\mathrm{Hz}$ for a 7-DOF robot.
Second, algorithms proposed in the previous works are often limited by singularities where multiple solutions, that explain all sensor measurements, exist. In fact, such singularities may appear quite frequently when only JTS is considered, as analyzed in \cite{pang2021identifying, popov2021real}. 

This paper proposes a solution that tackles the two challenges mentioned above. First, inspired by \cite{popov2021real}, we improve the run-time of PF with a mesh preprocessing that computes mesh data offline; e.g., distance between all face pairs. In our setup, however, this can only be done for each link separately, and consequently, particles cannot be updated across links. Therefore, we modified the PF by introducing \textit{exploration particles}, which are forced to investigate adjacent links. Second, we mitigate the singularity by introducing a F/T sensor at the base of the robot as shown in Fig. \ref{fig:vector_convention}(a). 
With this setup, we can estimate a contact at any link including the link 1 and 2, and can estimate multiple contacts even for a collaborative robot which has only a few DOF. We remark that these are known to be very challenging when the robot is equipped with only JTSs \cite{pang2021identifying, popov2021real}. 

To summarize, this paper tackles a problem of estimating contact forces and locations under multi-contact scenario for a collaborative robot arm. We propose a new algorithm called Multi-Contact Particle Filter with Exploration Particles (MCP-EP) which employs CPF (contact particle filter) \cite{manuelli2016localizing} as a backbone. Again, compared to the prior work, (i) the run-time is improved by introducing mesh preprocessing step and \textit{exploration particles}, and (ii) the base F/T sensor is used to tackle multi-contact scenario while mitigating the singularity.
With the proposed method, for a 7-DOF collaborative robot arm, the computation time was 0.45$\mathrm{ms}$ (2200$\mathrm{Hz}$) for a single-contact, and 1.66$\mathrm{ms}$ (600$\mathrm{Hz}$) for a dual-contact. Success rate of 99.96\% for a single-contact, could be achieved by alleviating singularities using the base F/T sensor. Multi-contact scenarios up to 3 contacts are also reported.

This paper is organized as follows. Section \ref{sec:contact_force_estimation} presents a QP formulation for the contact force identification that incorporate JTS and base F/T sensor. In section \ref{sec:particle_filter}, MCP-EP for multi-contact point localization and force identification is presented. In section \ref{sec:experiment}, the proposed algorithm is validated through simulation studies.

\section{QP-based Contact Force Identification without Localization}
\label{sec:contact_force_estimation}

This section presents a QP formulation that estimates contact forces, without contact point localization which will be presented in Section \ref{sec:particle_filter}. The QP outputs the contact forces as a decision variable that explains the sensor measurements (JTS and base F/T sensor) at a given candidate contact point. Before formulating the QP problem, we first present an estimation of external joint torque and contact-induced base wrench using proprioceptive sensors (i.e. JTS and base 6-axis F/T sensor). Throughout the paper, we assume a point contact that does not include a moment \cite{8059840, zwiener2018contact, zwiener2019armcl, bimbo2019collision ,popov2019real, likar2014external, manuelli2016localizing}. 

\subsection{Estimating External Joint Torque using JTS}
\label{subsec:jts}
Consider the following articulated multi-body dynamics (also known as link dynamics of a flexible joint robot model):

\begin{align}
\label{eq:robot_dyn}
\boldsymbol{M(q)\ddot{q}+C(q,\dot{q})\dot{q}+g(q)} = \boldsymbol{\tau}_{j} + \boldsymbol{\tau}_{ext},
\end{align}
where $\boldsymbol{q}\in\mathbb{R}^{n}$ is the joint variable,  $\boldsymbol{M(q)}$ is the link-side inertia matrix, $\boldsymbol{C(q,\dot{q})}$ is the Coriolis/centrifugal matrix, $\boldsymbol{g(q)}$ is the gravity vector, $\boldsymbol{\tau}_{j}$ is the joint torque measurable using JTS, and $\boldsymbol{\tau}_{ext}$ is the external torque caused by contact forces. 

Suppose that $k$ contacts exist in the robot (up to one contact per link), as shown in Fig. \ref{fig:vector_convention}(a). The external joint torque $\tau_{ext}$ is
\begin{align}
\label{eq:external_torque}
\boldsymbol{\tau}_{ext} = \sum_{i=1}^{k} \boldsymbol{J}_{i}^{T}(\boldsymbol{q}, \boldsymbol{r}_{c,i}) \boldsymbol{F}_{ext,i},
\end{align}
where $\boldsymbol{r}_{c,i} \in \mathbb{R}^{3}$ is the position vector of a $i^{\text{th}}$ contact point and $\boldsymbol{J}_{i}(\boldsymbol{q}, \boldsymbol{r}_{c,i}) \in \mathbb{R}^{3 \times n}$ is the associated positional Jacobian matrix at the contact point.
For a robot equipped with JTS, $\boldsymbol{\tau}_{ext}$ can be estimated using the following momentum-based observer \cite{de2006collision,kim2015design}:
\begin{align}
\label{eq:DOB}
&\begin{aligned}
\boldsymbol {\hat{\tau}}_{ext}(t) &= \mathbf{K}_{\text{o}}\left\{\boldsymbol {p}-\int _0^t 	\left(\boldsymbol {\tau }_{j} + \boldsymbol{n(q,\dot{q})} +\boldsymbol {\hat{\tau}}_{ext}\right) \textrm {d}s\right\},
\end{aligned}
\end{align}
\noindent where $\boldsymbol p = \boldsymbol{M(q)\dot{q}}$ is the robot generalized momentum, $\boldsymbol{n(q,\dot{q})} = \boldsymbol{C}^{T}(\boldsymbol{q,\dot{q}})\boldsymbol{\dot{q}} - \boldsymbol{g(q)}$ and $\mathbf{K}_{\text{o}}$ is a diagonal gain matrix. This can be regarded as a first-order low pass filter of the actual external joint torque $\boldsymbol{\tau}_{ext}$. 
With sufficiently large $\mathbf{K}_{\text{o}}$, we can simply assume $\hat{\boldsymbol{\tau}}_{ext} \simeq \boldsymbol{\tau}_{ext}$ \cite{de2006collision},\cite{haddadin2008collision}.

\subsection{Estimating Wrench caused by Contact Forces using Base F/T Sensor}
\label{subsec:base_F/T}
Let ${\boldsymbol{W}}_{ext,b} \in \mathbb{R}^{6}$ denote the wrench at the base caused by the external contact forces. Then,
\begin{align}
\label{eq:base_wrench}
{\boldsymbol{W}}_{ext,b} = \sum_{i=1}^{k} \underbrace{\begin{bmatrix} \mathbf{I}_{3\times3}  \\ skew(\boldsymbol{r}_{c,i}) \end{bmatrix}}_{\triangleq \boldsymbol{X}_{c,i}(\boldsymbol{r}_{c,i})} \boldsymbol{F}_{ext,i} ,
\end{align}
 \noindent $skew(\cdot)$ makes the $\mathbb{R}^3$ vector a $\mathbb{R}^{3 \times 3}$ skew-symmetric matrix. In fact, $skew(\cdot)$ defines a cross product, i.e., $\forall a,b\in \mathbb{R}^3$
\begin{align}
\label{eq:cross_product}
skew(a)b = a \times b .
\end{align}

 It should be noted that the base F/T sensor measurement does not provide a wrench due to the contact forces directly, because the dynamics of the robot are included in the measurement. Therefore, we employ a method proposed in \cite{buondonno2016combining} to estimate $\boldsymbol{W}_{ext,b}$ using the base F/T sensor.
 First, compute a nominal $\boldsymbol{\ddot{q}}_n$ by subtracting (\ref{eq:DOB}) from the (\ref{eq:robot_dyn}) and inverting the inertia matrix. This nominal value represents a situation in which there are no contact forces. Second, given the robot state $(\boldsymbol{q,\dot{q}})$ and $\boldsymbol{\ddot{q}}_n$, use the recursive Newton-Euler Algorithm (RNEA) to compute the nominal base wrench which the base F/T sensor exerts on the robot base link. Then, we can obtain $\hat{\boldsymbol{W}}_{ext,b}$, which is an estimate of ${\boldsymbol{W}}_{ext,b}$, by subtracting the nominal base wrench from the base F/T sensor measurement.

\subsection{QP Formulation for Contact Force Identification}
\label{subsec:QPformulation}

Define the vector $\hat{\boldsymbol{W}}$ by 
\begin{align}
    \hat{\boldsymbol{W}} = \begin{bmatrix} \hat{\boldsymbol{\tau}}_{ext} \\ \hat{\boldsymbol{W}}_{ext,b} \end{bmatrix} \in \mathbb{R}^{n+6}.
\end{align}
Now, the problem is to find the contact locations and forces that best explain $\hat{\boldsymbol{W}}$. 
Since we are assuming a point contact, a pulling force cannot be applied on the robot surface. To mathematically express this, we constrain the $i^{\text{th}}$ contact force to lie inside a friction cone $\mathcal{F}_{c}(\boldsymbol{r}_{c,i})$, i.e., $\boldsymbol{F}_{c,i} \in \mathcal{F}(\boldsymbol{r}_{c,i})$. 

From (\ref{eq:external_torque}), (\ref{eq:base_wrench}), and the friction cone constraint, we can formulate an optimization problem in which the contact points and forces are decision variables:
\begin{equation}
\centering
\begin{gathered}
\label{eq:non-convex}
    \displaystyle \min_{\boldsymbol{r}_c , \boldsymbol{F}_c}  \left \| \hat{\boldsymbol{W}} - \sum_{i=1}^{k} \begin{bmatrix} \boldsymbol{J}_{i}&  \boldsymbol{X}_{c,i} ^T  \end{bmatrix}^T \boldsymbol{F}_{c,i}   \right \|^{2} \\
    \text{subject to} \,\,\, \boldsymbol{r}_{c,i} \in \mathcal{S}, \boldsymbol{F}_{c,i} \in \mathcal{F}(\boldsymbol{r}_{c,i}), \;\;  \forall i \in \{1,\ldots,k\},
\end{gathered}
\end{equation}

\noindent where $\boldsymbol{r}_c = (\boldsymbol{r}_{c,1}, \ldots, \boldsymbol{r}_{c,k} )^{T}$, $\boldsymbol{F}_c = (\boldsymbol{F}_{c,1}, \ldots, \boldsymbol{F}_{c,k})^{T}$ and $\mathcal{S}$ represents the robot surface.
This is a nonlinear optimization problem since, for example, $\boldsymbol{r}_c$ and $\boldsymbol{F}_c$ appear as a cross product in $\begin{bmatrix} \boldsymbol{J}_{i}&  \boldsymbol{X}_{c,i} ^T  \end{bmatrix}^T \boldsymbol{F}_{c,i}$ as can be seen in (\ref{eq:base_wrench}) and (\ref{eq:cross_product}).

To simplify this problem, the friction cone is approximated as a polyhedral cone as shown in Fig. \ref{fig:vector_convention}(b). Under this  approximation, the $i^{\text{th}}$ contact force can be expressed as 
\begin{align}
\label{eq:friction_cone}
\boldsymbol{F}_{c,i}=\begin{bmatrix}\boldsymbol{a}_{c,i,1} & \ldots & \boldsymbol{a}_{c,i,4} \end{bmatrix} \begin{bmatrix} f_{c,i,1}\\ \vdots \\f_{c,i,4} \end{bmatrix} = \boldsymbol{A}_{c,i} \boldsymbol{f}_{c,i},
 \end{align}
where $\boldsymbol{a}_{c,i,j}\in\mathbb{R}^{3}$ is a support vector of the polyhedral friction cone and $f_{c,i,j}$ is a weight of each support vector. 

For simplicity, in (\ref{eq:non-convex}), let us express $\small{\sum_{i=1}^{k} \begin{bmatrix} \boldsymbol{J}_{i}&  \boldsymbol{X}_{c,i} ^T  \end{bmatrix}^T \boldsymbol{F}_{c,i}}$ as $\boldsymbol{Q}^{T}\boldsymbol{f_{c}}$ with $\boldsymbol{f}_c=(\boldsymbol{f}_{c,1},\ldots,\boldsymbol{f}_{c,k})^{T}$. Here, $\boldsymbol{Q} \in \mathbb{R}^{4k \times (n+6)}$ is a proper augmentation of $\boldsymbol{A}_{c,i}^T \boldsymbol{J}_{i}$ and $\boldsymbol{X}_{c,i} \boldsymbol{A}_{c,i}$ matrices. If, in addition, $\boldsymbol{r}_{c}$ is fixed, then (\ref{eq:non-convex}) can be written as
\begin{equation}
\begin{gathered}
    \label{eq:convex}
    \displaystyle \min_{\boldsymbol{f}_c}  \left \| \hat{\boldsymbol{W}} -  \boldsymbol{Q}^T\boldsymbol{f}_c \right \|^{2} \\
    \text{subject to} \,\,\, f_{c,i,1},\ldots,f_{c,i,4} \leq 0, \;\;  \forall i \in \{1,\ldots,k\},
\end{gathered}
\end{equation}
\noindent which is a convex QP problem because $\boldsymbol{Q}$ is constant for the fixed $\boldsymbol{r}_c$ and the friction cone is a convex set. Note that the contact force $\boldsymbol{f}_c$ is the only decision variable for this problem.

It is worthwhile to mention that, by virtue of the base F/T sensor, 6 sensing DOFs are added to (\ref{eq:convex}) regardless of the contact location. For example, if a contact is in link 1, the number of available sensor DOF is 7 (6 from the base F/T sensor and 1 from the JTS at joint 1). In contrast, if the robot is equipped with only JTSs, the sensing DOF, in this case, is 1, and therefore, the contact force and location cannot be estimated.

%


\section{Particle filter Algorithm for Contact Point Localization and Force Identification}
\label{sec:particle_filter}

This section presents the MCP-EP algorithm for multi-contact point localization and force identification. In this section, $\mathcal{X}_{i}$ represents a $i^{\text{th}}$ set of particles $\{x_i^{[1]}, \cdots, x_i^{[m]}\}$, and $\mathfrak{X} = \{\mathcal{X}_{1},\ldots,\mathcal{X}_{k} \}$ represents a set of particle sets. The weight of a particle $x_i^{[m]}$ is denoted by $w_i^{[m]}$. We assume that multi-contact occurs sequentially, which is reasonable from the practical point of view. The $i^{\text{th}}$ particle set $\mathcal{X}_{i} \in \mathfrak{X}$ will estimate the $i^{\text{th}}$ contact point.

\subsection{Motion-Model}
\label{subsec:motion_model}

\begin{figure}[tb!]
\centering
    \subfigure[]
	{\includegraphics[width=0.22\textwidth]{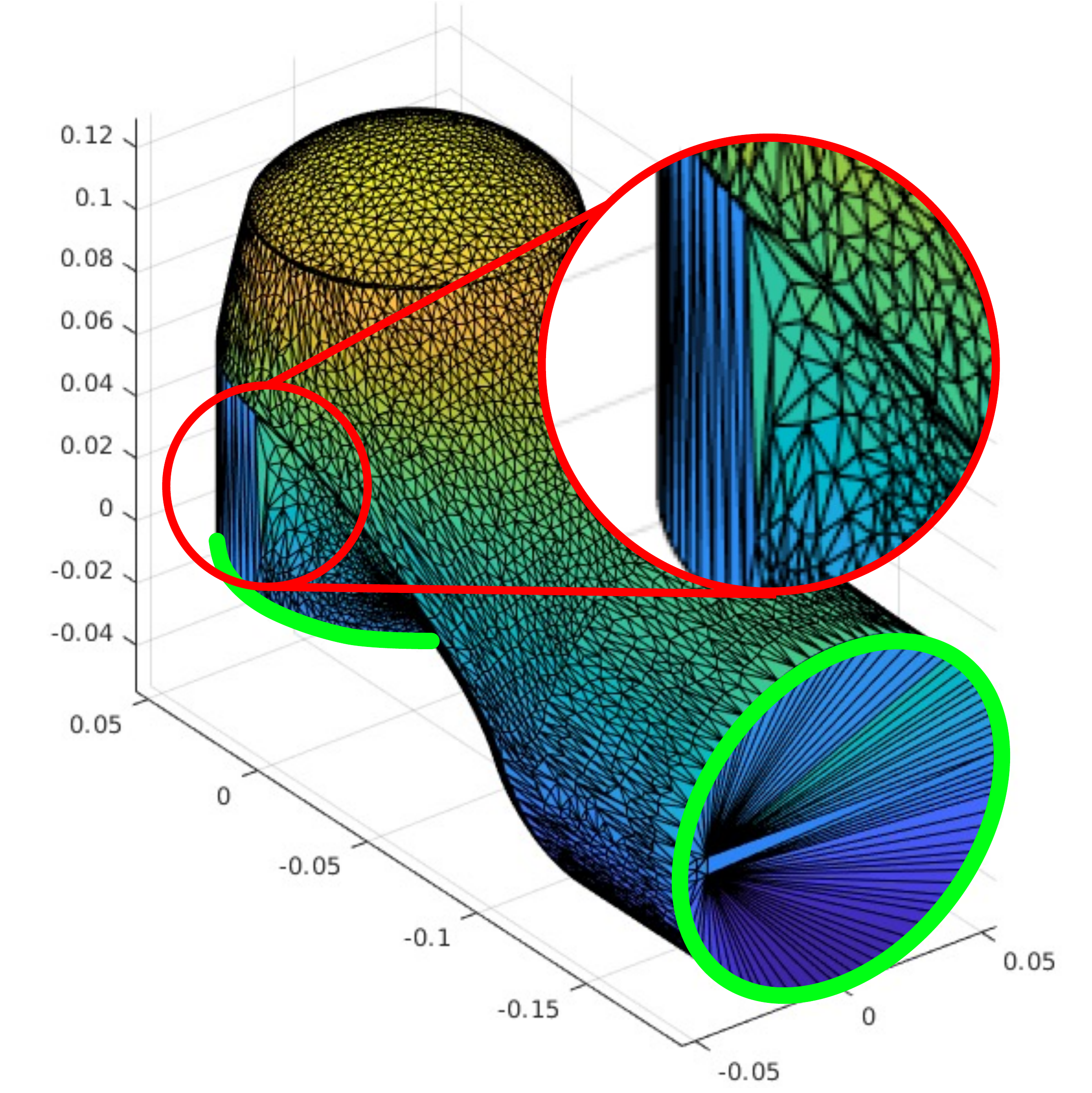}}
	\subfigure[]
	{\includegraphics[width=0.22\textwidth]{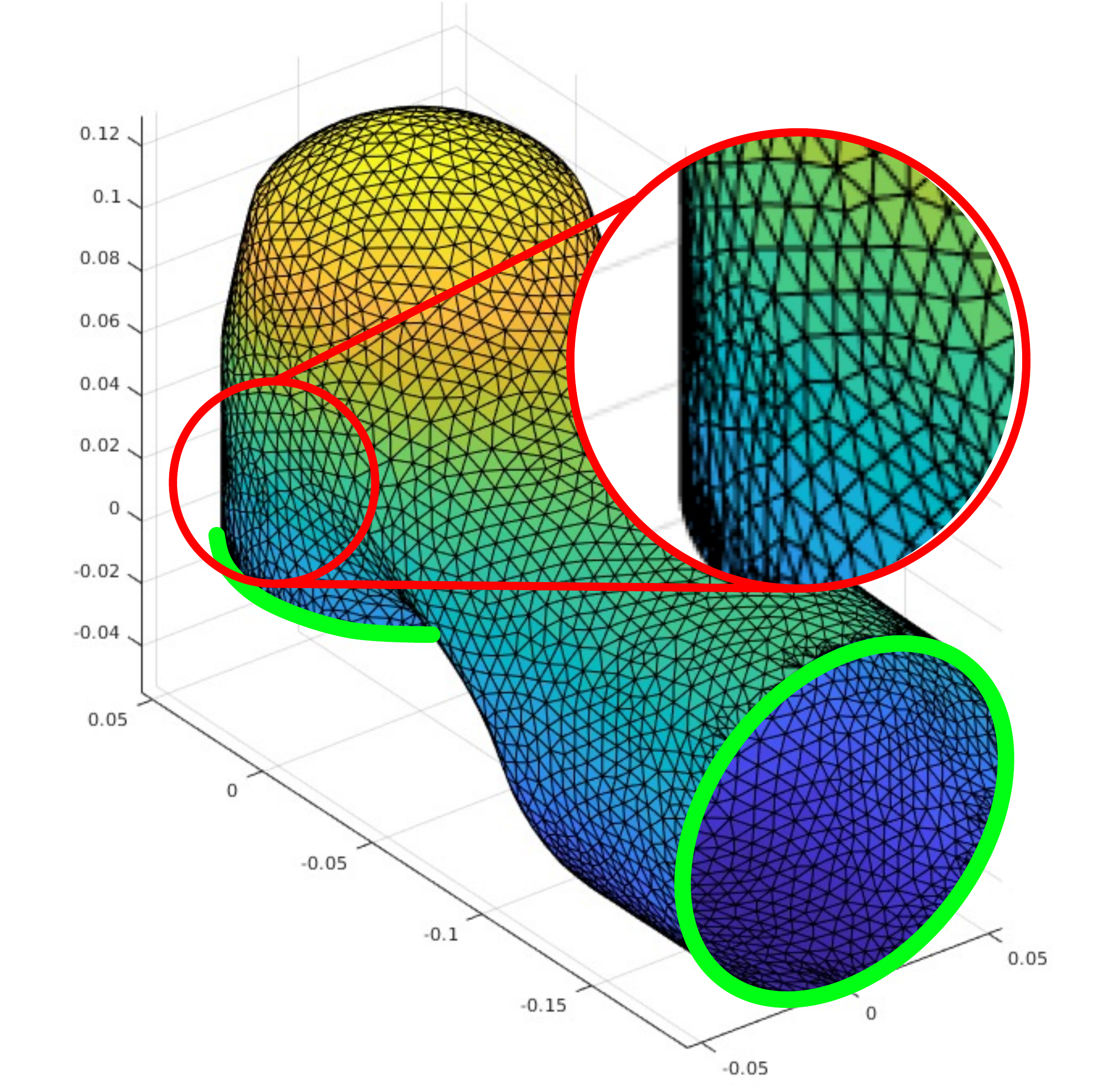}}
	\subfigure[]
	{\includegraphics[width=0.2\textwidth]{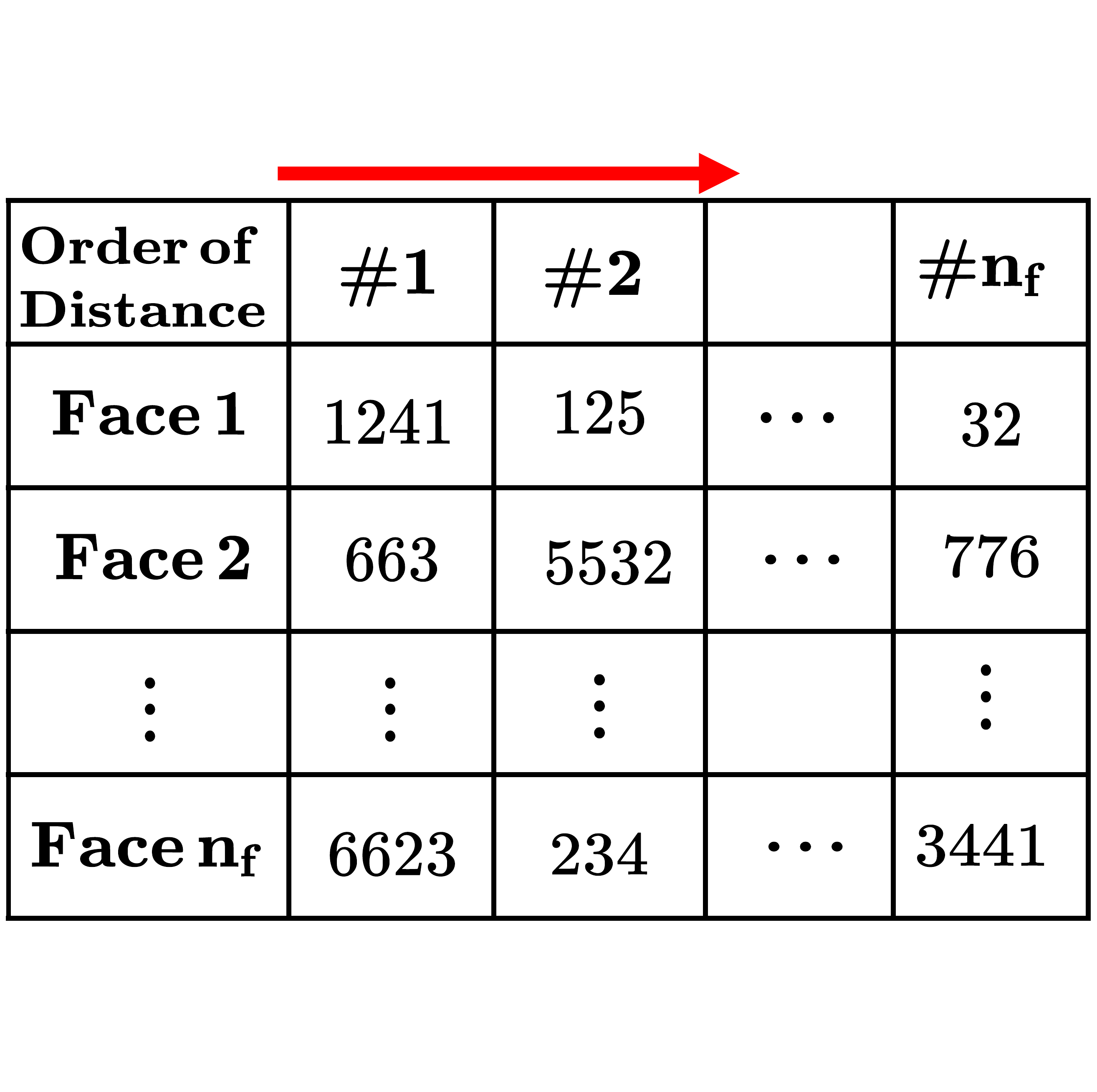}}\;\;\;\;
	\subfigure[]
	{\includegraphics[width=0.17\textwidth]{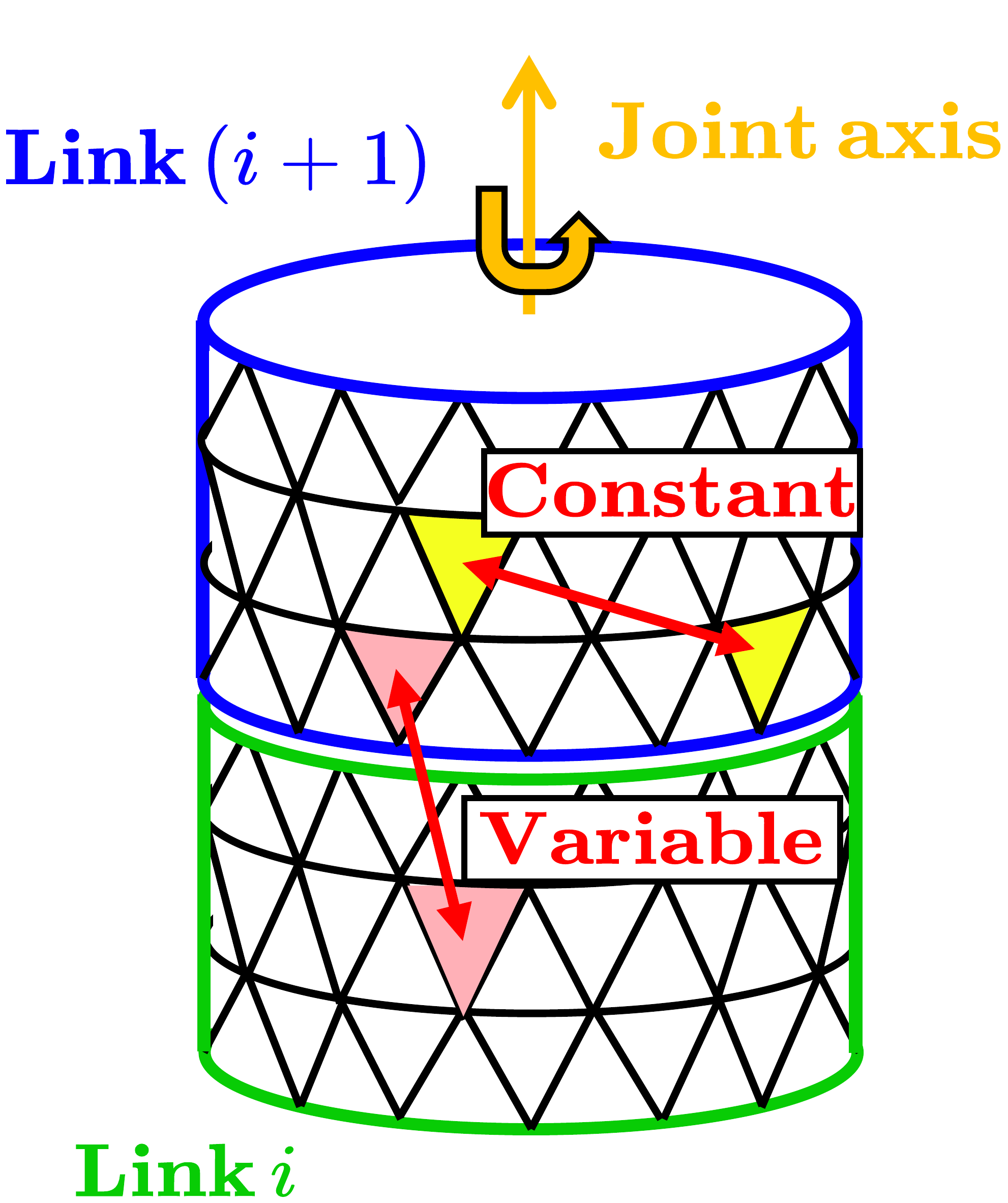}}
	
	\caption{
	(a) The original mesh has irregular triangle faces (see red circle) and non-contact parts (see green circle) 
	(b) Non-contact parts are eliminated, and the size of the faces is almost uniform. 
	(c) The data structure contains adjacent face indices in ascending order of the geodesic distance.
	(d) While the distance between the yellow faces is constant, the distance between red faces changes as the joint rotates. Namely, the distance information between different links cannot be precomputed.
	}
	\label{fig:processing_mesh}
	\vspace{-5mm}
\end{figure}

When updating particles according to the motion model of PF, as there is no input to the motion model, a common way is to update particles by a random walk method \cite{manuelli2016localizing,zwiener2019armcl}. 
To improve the algorithm run-time, in this paper, we perform a mesh preprocessing, and modify the motion model accordingly.

\subsubsection{Mesh preprocessing}
\label{subsubsec:preprocessing}

As shown in the green circle of Fig. \ref{fig:processing_mesh}(a), there exists a region in which a contact cannot occur. We removed this manually using Blender \cite{Hess:2010:BFE:1893021}. Moreover, since the original mesh is irregular (red circle in Fig. \ref{fig:processing_mesh}(a)), we additionally applied a isotropic remesh algorithm \cite{yan2009isotropic}. Fig. \ref{fig:processing_mesh}(b) shows the resulting faces. As a result, we can represent the position of a particle using the link and face index number as follows:
\begin{equation}
    x_i^{[m]}=(link\#, \; face\#). 
    \label{eq:particle_definition}
    \vspace{-2.5mm}
\end{equation}
Moreover, we compute the geodesic distances (the shortest distance on the surface) for all face pairs using GP-toolbox \cite{gptoolbox} and stored them in ascending order as shown in Fig. \ref{fig:processing_mesh}(c). With this data structure, we can quickly find nearby faces at any particle. Note that the whole procedure is computed offline, imposing no computation burden on the MCP-EP algorithm.
\subsubsection{Modified Motion Model}
\label{subsubsec:motion_model}
Using the data structure in Fig. \ref{fig:processing_mesh}(c), the random walk can be applied simply by changing the face index of a particle (i.e $face\#$ of $x_i^{[m]}$). Since no algorithmic computation is required other than index changing, the computation cost is extremely low. For our implementation, it takes about 0.06$\mu$s to update one particle. 

However, as shown in the Fig. \ref{fig:processing_mesh}(d), the distance between the faces in different links varies as the joint rotates. Therefore, the pre-calculated table in Fig. \ref{fig:processing_mesh}(c) is valid only for each link, and consequently, the motion model is not able to update particles across the link; i.e., the motion model changes $face\#$, but not $link\#$. To overcome this, we will introduce \textit{exploration particles} later in Section \ref{subsec:exploration_particle}.
\begin{figure}[tb!]
    \centering
    \subfigure[]
	{\includegraphics[width=0.235\textwidth]{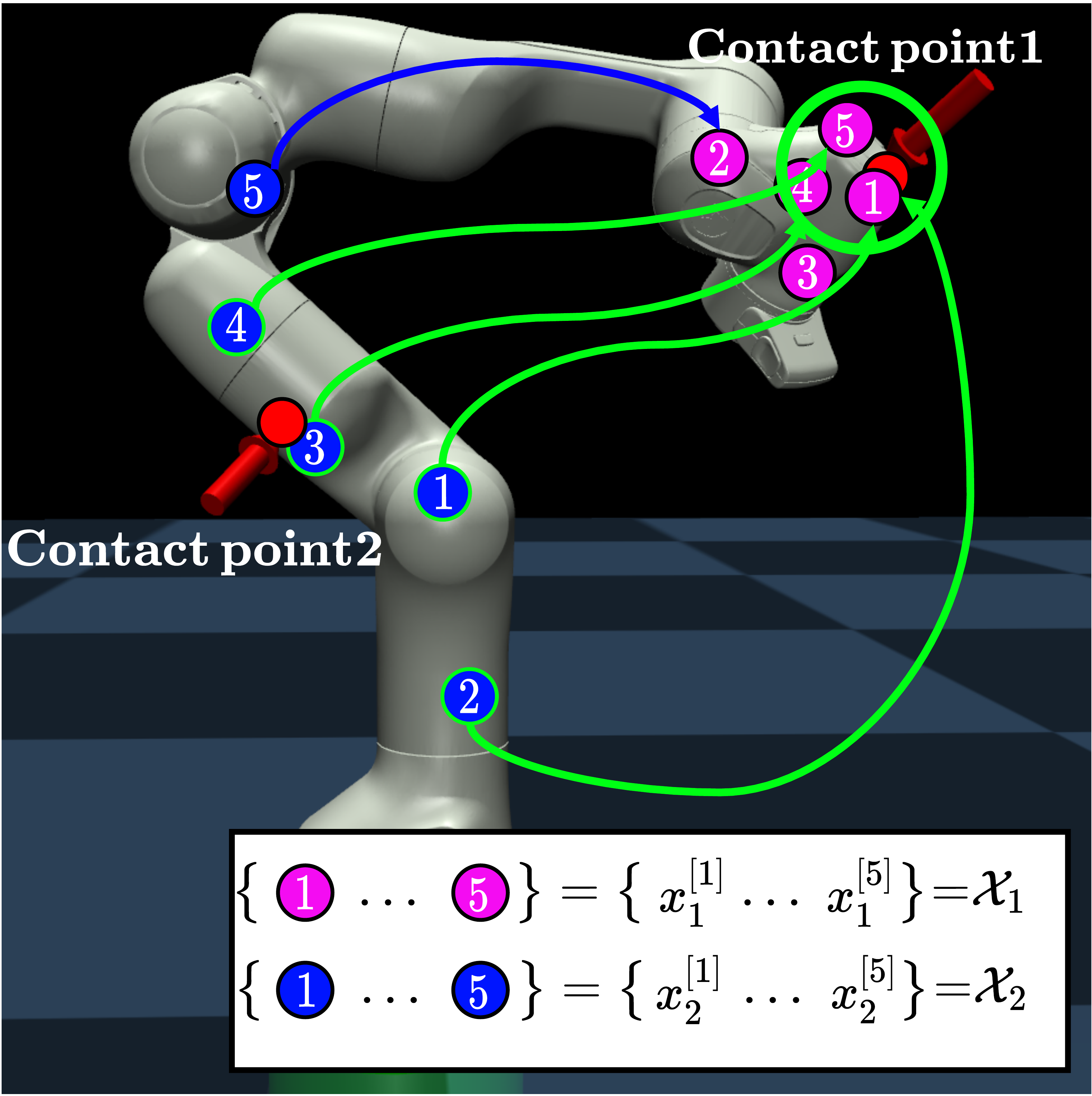}}
	\subfigure[]
	{\includegraphics[width=0.235\textwidth]{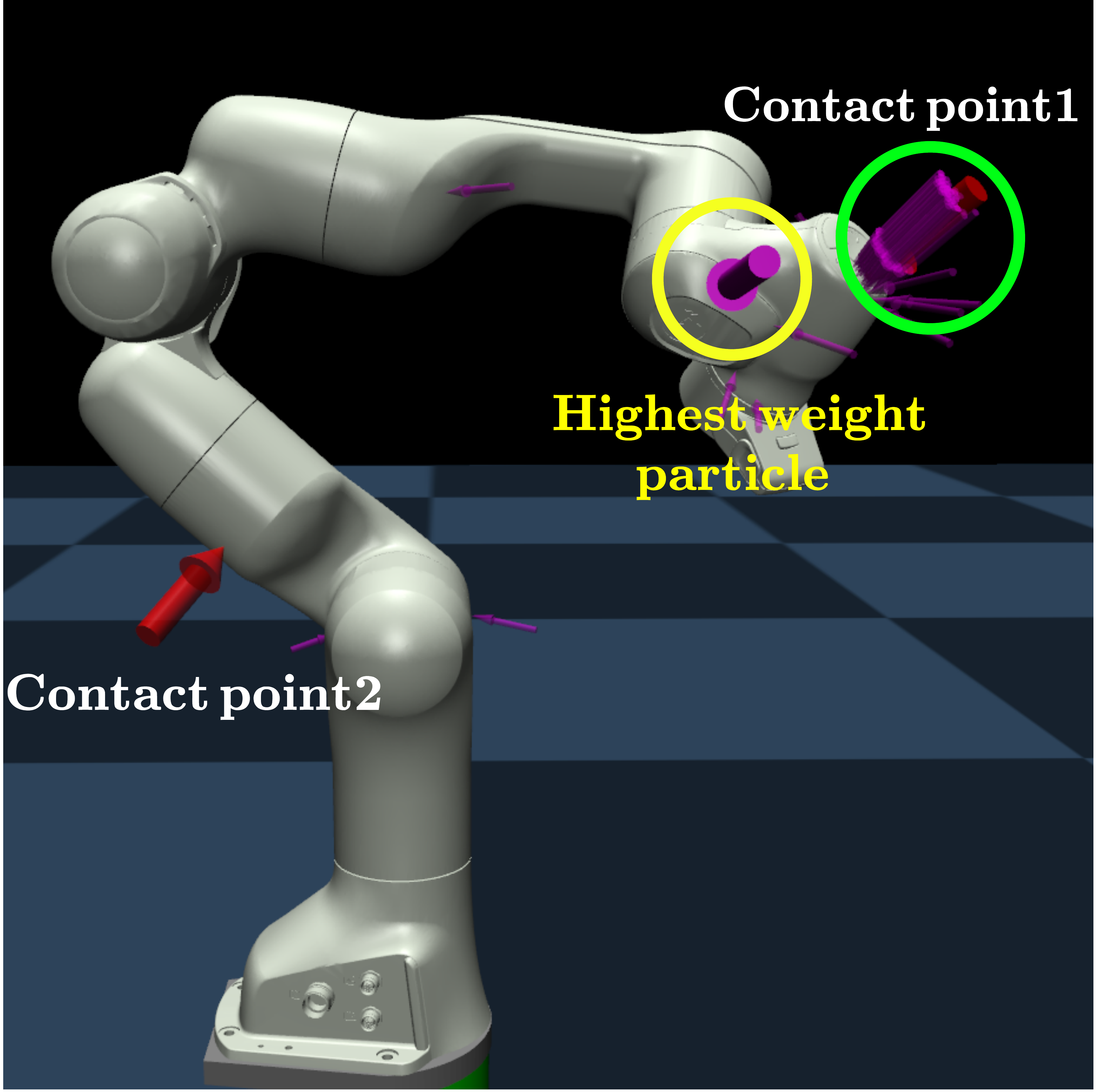}}
    \caption{
    (a) Since $\mathcal{X}_1$ is already converged to the contact point 1, a weight of particle in $\mathcal{X}_2$ can be updated with sampled particle ($x^*(\mathcal{X}_{1})$).
    (b) Suppose that the second force is applied, but a new particle set is not added yet. The particle with the highest weight and the true contact point 1 are indicated by the yellow and green circles, respectively. In this case, the particle with the highest weight does not represent the true contact point 1.
    }
    \label{fig:multi_weight_update}
    \vspace{-5mm}
\end{figure}
\subsection{Measurement-Model}
\label{subsec:measurement_model}

Since a smaller QP error implies that the particle well-explains the sensor measurements, we can define the weight of a particle $w_i^{[m]}$ using the QP error. Namely, a particle with a smaller QP error is more likely a true contact point. For a single-contact case, therefore, we can compute associated weights by evaluating the QP error for all particles.

The multi-contact case, however, becomes more complicated, as there exist multiple particle sets. When computing the weight of a particle, other particle sets should be considered jointly. This can also be captured by the fact that the $\boldsymbol{Q}$ matrix in the QP problem (\ref{eq:convex}) has the size of $4k\times(n+6)$, where $k$ is the number of contacts. Since evaluating all possible pairs of particles will require a huge amount of computation, we propose a sampling-based approach as follows.

Suppose that the weight $w_{1}^{[m]}$ of a particle $x_{1}^{[m]}\in\mathcal{X}_{1}$, among $k>1$ particle sets, is of interest. Then, let us sample particles from the other particle sets: 
$\boldsymbol{x}^*=\{x^*(\mathcal{X}_2),\; x^*(\mathcal{X}_3),\; \ldots ,\; x^*(\mathcal{X}_{k}) \}$, where $x^*(\mathcal{X}_j)$ represents a sampled particle from $\mathcal{X}_{j}$ considering the associate weights.
With this setup, the weight $w_{i}^{[m]}$ can be computed by evaluating the following QP problem:
\begin{align}
\label{eq:weight}
      w_i^{[m]} = p(\hat{\boldsymbol{W}}\mid x_i^{[m]}) \propto \mathbf{exp}\left ( -\alpha QP(\hat{\boldsymbol{W}}\mid x_i^{[m]}, \boldsymbol{x}^{*})\right ),
\end{align}
\begin{align}
\label{eq:QP}
        QP(\hat{\boldsymbol{W}}\mid x_i^{[m]}, \boldsymbol{x}^*) = \displaystyle \min_{\boldsymbol{f}_c}  \left \| \hat{\boldsymbol{W}} -  \boldsymbol{Q}^T\boldsymbol{f}_c \right \|^{2},
\end{align}

\noindent where $\alpha > 0$ to ensure that the weight is inversely proportional to the QP error.

Updating the weight of a particle in this way assumes that the other particle sets have already converged to the true contact points. This assumption is reasonable because forces are sequentially applied to the robot, while the proposed algorithm is fast enough to converge before the next contact arrives. For example, in Fig. \ref{fig:multi_weight_update}(a), the weight of a particle in $\mathcal{X}_2$ can be updated with a sampled particle $x^*(\mathcal{X}_{1})$ which can estimate the true contact point quite reasonably.

\begin{rem}
When a new contact occurs, the algorithm adds a new particle set, as will be presented later. However, the algorithm is not able to recognize it immediately, and in the meantime, the particles tend to move around to explain the new contact without an additional particle set, as shown in Fig. \ref{fig:multi_weight_update}(b). In our experience, sampling $\boldsymbol{x}^*$ showed more robustness than manually selecting $\boldsymbol{x}^*$ with a certain criterion (e.g. selecting a particle with the highest weight) because sampling gives randomness to the measurement model. 
For example, in the case of Fig. \ref{fig:multi_weight_update}(b), the particle in the yellow circle does not represent the true contact point (in the green circle), despite the fact that it has the highest weight.

\end{rem}

\begin{figure}[tb!]
\centering
    \subfigure[]
	{\includegraphics[width=0.235\textwidth]{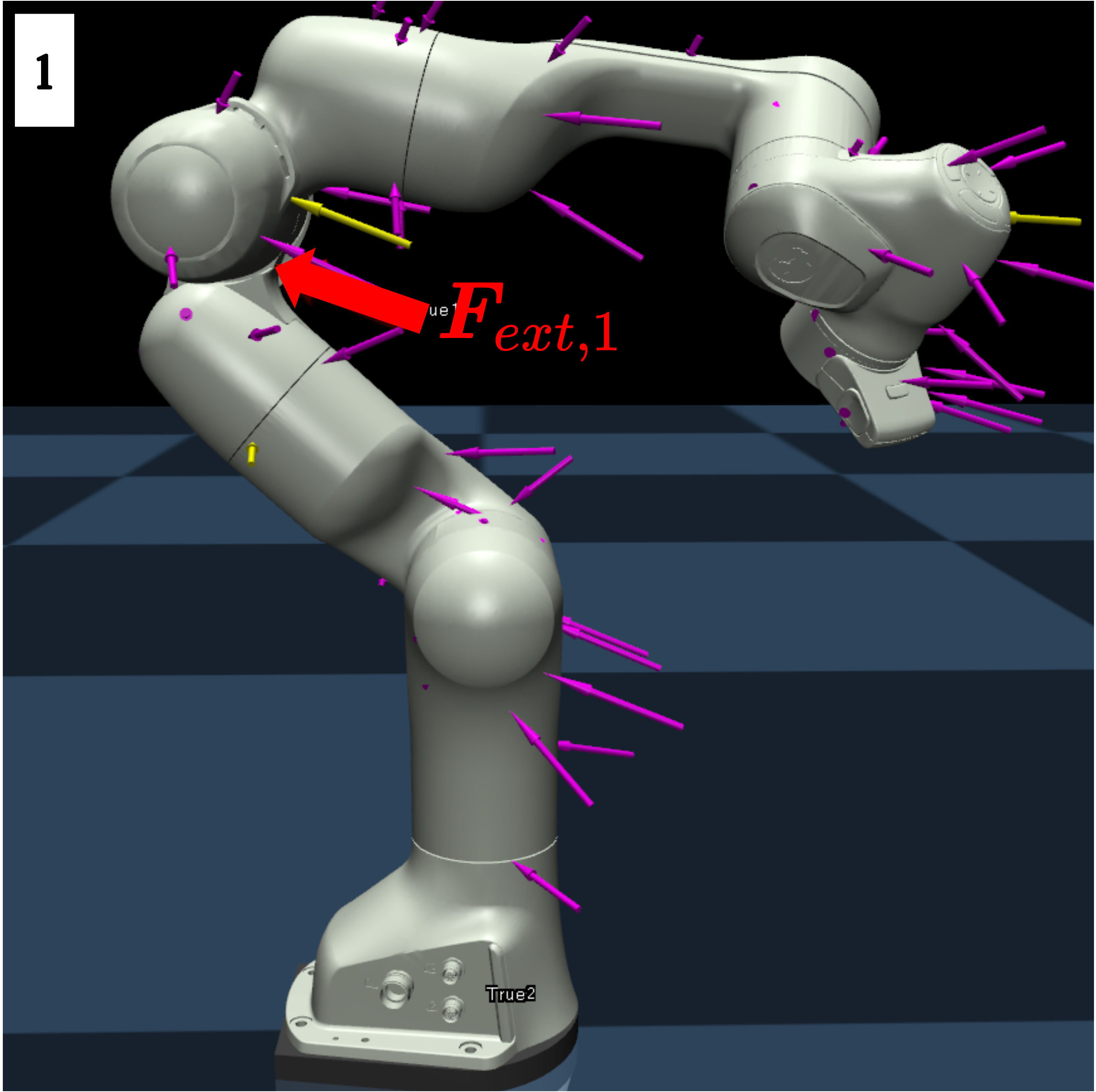}}
	\subfigure[]
	{\includegraphics[width=0.235\textwidth]{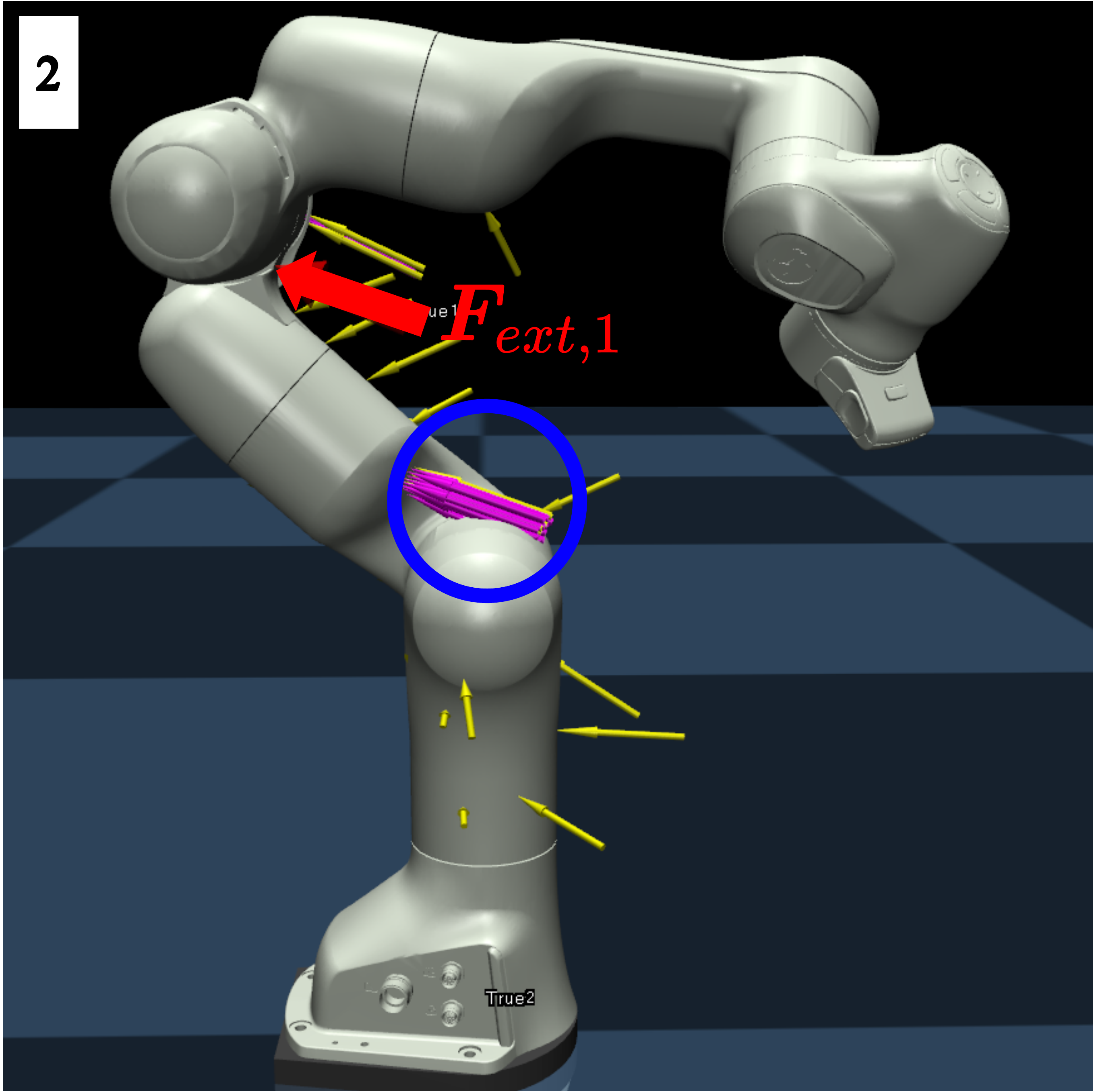}}
	\subfigure[]
	{\includegraphics[width=0.235\textwidth]{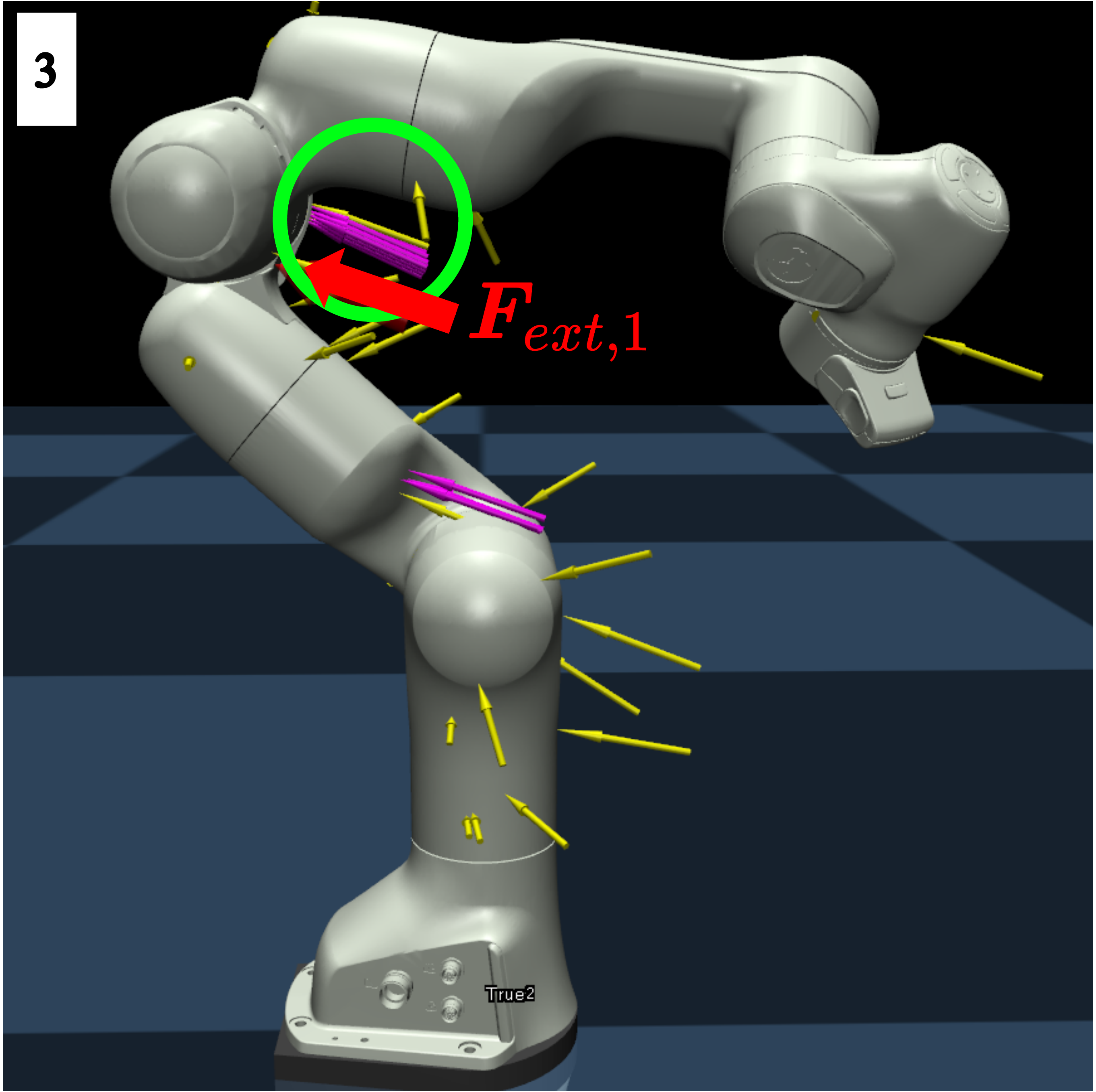}}
	\subfigure[]
	{\includegraphics[width=0.235\textwidth]{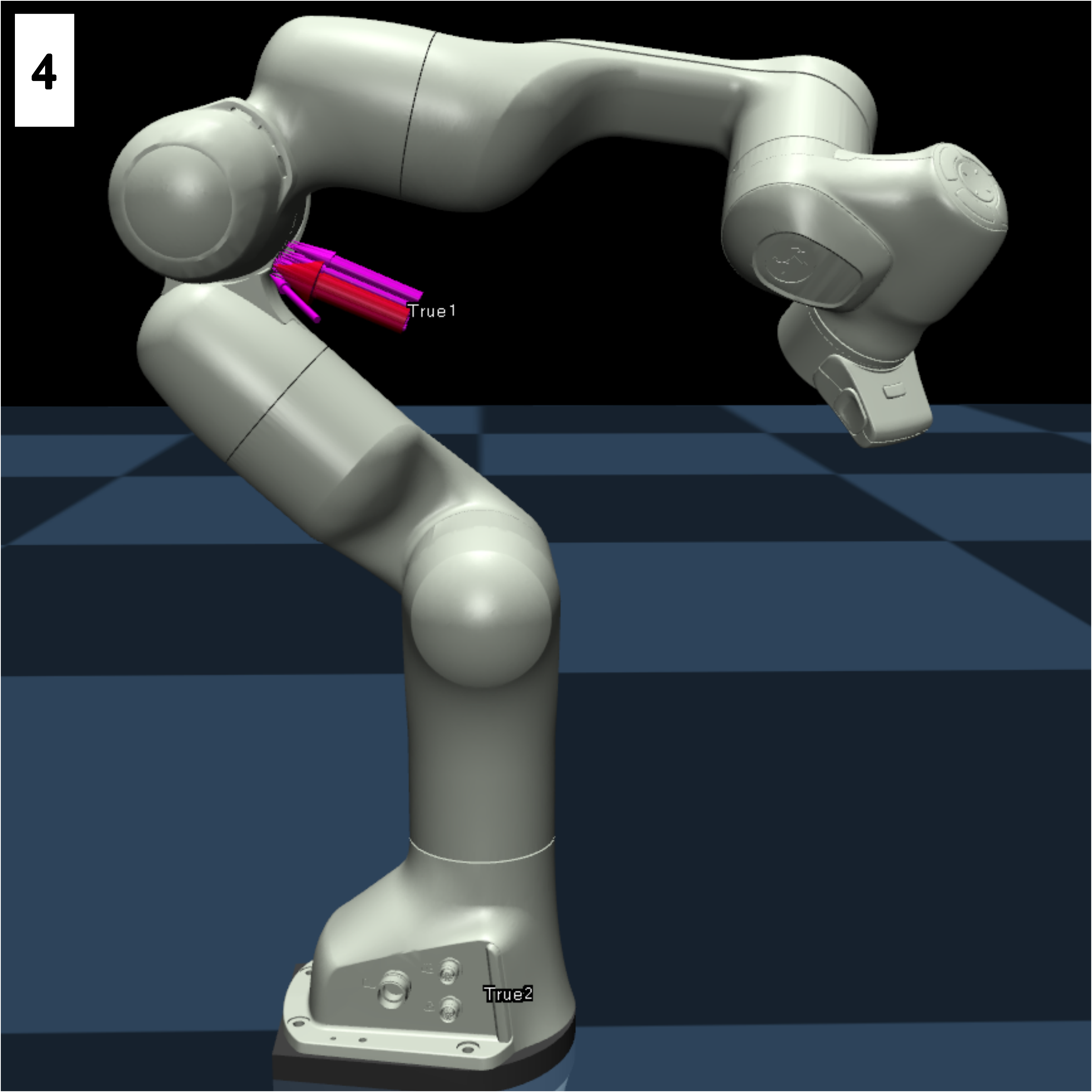}}
	\caption{(a) A contact force is applied to link 3, and a particle set is initialized and scattered over the entire robot. 
    (b) Pink and yellow arrows indicate the basic and exploration particles, respectively. All basic particles converged to link 2 (see the blue circle). Since basic particles do not well-explain sensor measurements in the sense of QP error, exploration particles are scattered into the adjacent links.
	(c) Some basic particles are resampled around the exploration particles which better explain the sensor measurements (see the green circle).
	(d) Eventually, all basic particles converge to the true contact point in link 3. 
	}
	\label{fig:exploration_particle}
	\vspace{-5mm}
\end{figure}

\subsection{Exploration Particle}
\label{subsec:exploration_particle}

Recall that, the motion model is not able to update particles across the links because it only changes $face\#$. Therefore, once all particles are converged to a wrong link, there is no chance that particles are resampled in a correct link. To overcome this, we introduce \textit{exploration particles} to investigate adjacent links (see Fig. \ref{fig:exploration_particle}(a) and (b)). 

\begin{algorithm}[htb!]
	\caption{Update-Exploration-Particles($\mathfrak{X}$)}
    \begin{algorithmic}[1]
    \For {$\mathcal{X}_i$ in $\mathfrak{X}$}
	\If {Get-Maximum-Weight($\mathcal{X}_i) < \bar{\epsilon}$}
	    \State sample $x_{i}^{[m]}$ $\in\mathcal{X}_{i,basic}$
	    \State discard $x_{i}^{[m]}$ from $\mathcal{X}_{i,basic}$
	    \State $x_{i}^{[m]} \leftarrow$ Motion-Model-near-link($x_{i}^{[m]}$)
	    \State add $x_{i}^{[m]}$ to $\mathcal{X}_{i,explore}$
	\Else
	    \State $\mathcal{X}_{i,basic} = \mathcal{X}_{i,basic} \cup \mathcal{X}_{i,explore}$
	    \State $\mathcal{X}_{i,explore}= \emptyset$
	\EndIf 
	\EndFor 
	\State \Return{$\mathfrak{X}$}
	\end{algorithmic}
	\label{alg:ep}
\end{algorithm}

\subsubsection{Update-Exploration-Particles}
\label{subsubsec:update_ep}
Algorithm \ref{alg:ep} introduces a method for updating the \textit{exploration particles}. 
The main idea is to divide the particle set $\mathcal{X}_{i}$ into two subsets: basic ($\mathcal{X}_{i,basic}$) and explore ($\mathcal{X}_{i,explore}$). 
If the maximum weight of the particle set is lower than a certain threshold $\bar{\epsilon}$ (i.e particle set does not well-explain the contact point),
one particle is randomly discarded from $\mathcal{X}_{i,basic}$. Then using Motion-Model-near-Link (line 5), a new particle with adjacent $link\#$ is added to $\mathcal{X}_{i,explore}$.
Otherwise, merge $\mathcal{X}_{i,explore}$ into $\mathcal{X}_{i,basic}$ (line 8-9).

\subsubsection{Importance-Resampling-EP (Exploration Particles)}
\label{subsubsec:resample_ep}
\textit{Exploration particles} undergo a different resampling process than the particles in $\mathcal{X}_{i,basic}$. Basic particles are updated in the same way as a standard importance resampling process. However, \textit{exploration particles} do not undergo resampling. That is, basic particles can be resampled around \textit{exploration particles}, but \textit{exploration particles} do not disappear during the resampling process.
Even if all basic particles converged to a wrong link, they can be resampled on the correct link due to the existence of \textit{exploration particles} (see Fig. \ref{fig:exploration_particle}(c) and (d)).

\begin{algorithm}[htb!]
	\caption{Manage-Particle-Set($\mathfrak{X}$)}
    \begin{algorithmic}[1]
	\If {$\epsilon (\hat{\boldsymbol{W}},\mathfrak{X}) > \bar{\epsilon}$}
	    \State add $\mathcal{X}_{init} $ to $\mathfrak{X}$
	    \State \Return{$\mathfrak{X}$}
	\EndIf
	\For {$\mathcal{X}_i$ in $\mathfrak{X}$}
	    \If{$\epsilon (\hat{\boldsymbol{W}},\mathfrak{X}\backslash \mathcal{X}_i ) < \bar{\epsilon}$}
	        \State discard $\mathcal{X}_i$ from $\mathfrak{X}$
	        \State \Return{$\mathfrak{X}$}
	    \EndIf
	\EndFor
	\State \Return{$\mathfrak{X}$}
	\end{algorithmic}
	\label{alg:manage}
\end{algorithm}
\subsection{Manage-Particle-Set}
\label{subsec:manage_particle_set}
Since multiple contacts occur sequentially, the number of particle sets should be properly adjusted. To this end, we adopt a method proposed in  \cite{manuelli2016localizing} without modification, which we present in Algorithm \ref{alg:manage} to make this paper self-contained.

In Algorithm \ref{alg:manage}, $\epsilon (\hat{\boldsymbol{W}},\mathfrak{X}) = QP(\hat{\boldsymbol{W}}\mid \boldsymbol{x}^{\#}(\mathfrak{X}))$, where $\boldsymbol{x}^{\#}(\mathfrak{X}) = \{{x}^{\#}(\mathcal{X}_{1}), \ldots , {x}^{\#}(\mathcal{X}_{k})\} $, and ${x}^{\#}(\mathcal{X}_{i})$ is a particle with the highest weight in $\mathcal{X}_i$. If $\mathfrak{X}$ is empty then define $QP(\hat{\boldsymbol{W}}\mid \boldsymbol{x}^{\#}(\mathfrak{X})) = \hat{\boldsymbol{W}}^T\hat{\boldsymbol{W}}$ to initialize the PF. In line 1-4, if the current particle set does not explain the sensor measurements, a new particle set $\mathcal{X}_{init}$ is added to $\mathfrak{X}$. In line 5-10, if sensor measurements can be well-explained without a certain particle set $\mathcal{X}_{i}$, discard it from $\mathfrak{X}$.

\begin{algorithm}[htb!]
   \caption{MCP-EP($\mathfrak{X}$)}
	\begin{algorithmic}[1]
	    \State $\mathfrak{X} \leftarrow$ Motion-Model$(\mathfrak{X})$                          \Comment{In Section \ref{subsec:motion_model}\;\;\,}
	    \State $\mathfrak{X} \leftarrow$ Measurement-Model$(\mathfrak{X})$                     \Comment{In Section \ref{subsec:measurement_model}\;\;\,}
	    \State $\mathfrak{X} \leftarrow$ Update-Exploration-Particles$(\mathfrak{X})$          \Comment{In Section \ref{subsubsec:update_ep}}
	    \State $\mathfrak{X} \leftarrow$ Importance-Resampling-EP$(\mathfrak{X})$              \Comment{In Section \ref{subsubsec:resample_ep}}
	    \State $\mathfrak{X} \leftarrow$ Manage-Particle-Set $(\mathfrak{X})$                  \Comment{In Section \ref{subsec:manage_particle_set}\;\;\,}
	    \State \Return $(\mathfrak{X})$
	\end{algorithmic}
	\label{alg:mcp-ep}
\end{algorithm}

\subsection{Obtaining Estimated Contact Point and Force from Particle Set}
\label{subsec:obtain_contact_point_force}
The overall procedure of MCP-EP is given in Algorithm \ref{alg:mcp-ep} in which aforementioned methods are executed sequentially. After iterating Algorithm \ref{alg:mcp-ep} several times, each particle set converges to each contact point.
The estimated $i^{\text{th}}$ contact point ($\boldsymbol{r}_{c,i}$) can be obtained using ${x}^{\#}(\mathcal{X}_i)$ which is the particle with the highest weight in $\mathcal{X}_i$. The estimated $i^{\text{th}}$ contact force ($\boldsymbol{F}_{c,i}$) can be obtained by solving the QP with given estimated contact points ($QP(\hat{\boldsymbol{W}}\mid \boldsymbol{x}^{\#}(\mathfrak{X}))$).

\section{Simulation Validation}
\label{sec:experiment}
The proposed MCP-EP algorithm is implemented in Mujoco simulator \cite{6386109}. Experiments are conducted with a 7-DOF Franka Emika Panda robot arm that is equipped with JTSs. The QP (\ref{eq:convex}) is solved using qpSWIFT \cite{pandala2019qpswift}. The algorithm was tested on Intel Core i7-12700 cpu, 32Gb RAM PC as a C++ single thread program.

Single, dual, and triple contact scenarios are simulated 10000 times each. At each run, robot configuration was initialized randomly, and 20$\mathrm{N}$ force in the friction cone was applied at a random position (however, up to 1 contact per link). For quantitative evaluation of the algorithm, we classify the case with $\leq$ 2.25$\mathrm{cm}$ (which corresponds to, roughly, 5 faces) localization error as success. Then, the success rate can be calculated as `\# of success/10000'. We also report convergence step, which means the number of iterations until succeeds, root mean square error (RMSE) of position and force, and the algorithm run-time (per iteration). Table \ref{tab:overview} summarizes the simulation results. Please notice that the RMSE is evaluated including fail cases. In the following, we discuss single and dual contact cases in more detail.

\begin{table}[tb!]
\resizebox{\columnwidth}{!}{%
\begin{tabular}{cccccccc}
\hline
\begin{tabular}[c]{@{}c@{}}\# of \\ contact\end{tabular} & \begin{tabular}[c]{@{}c@{}}\# of \\ particle\end{tabular} & \begin{tabular}[c]{@{}c@{}}Succ. \\ rate(\%)\end{tabular} & \begin{tabular}[c]{@{}c@{}}Conv.\\ step\end{tabular} & \begin{tabular}[c]{@{}c@{}}RMSE\\ position(cm)\end{tabular} & \begin{tabular}[c]{@{}c@{}}RMSE\\ force(N)\end{tabular} & \begin{tabular}[c]{@{}c@{}}Run-time\\ ($\mathrm{ms}$)\end{tabular} \\ \hhline{=======}
1  & 100 & 99.96 & 18.25 & 0.16 & 0.01 & 0.45 \\ \hline
2  & 200 & 96.70 & 11.00 & 1.08 & 2.00 & 1.66 \\ \hline
3  & 300 & 81.63 & 14.37 & 3.50 & 4.63 & 4.50 \\ \hline
\end{tabular}%
}
\caption{Simulation Result : Overview}
\label{tab:overview}
\vspace{-7mm}
\end{table}

\begin{table}[h]
\centering
\resizebox{\columnwidth}{!}{%
\begin{tabular}{clllllll}
\hline
Contact link & \multicolumn{1}{c}{\textbf{link1}} & \multicolumn{1}{c}{\textbf{link2}} & \multicolumn{1}{c}{\textbf{link3}} & \multicolumn{1}{c}{\textbf{link4}} & \multicolumn{1}{c}{\textbf{link5}} & \multicolumn{1}{c}{\textbf{link6}} & \multicolumn{1}{c}{\textbf{link7}} \\ \hhline{========}
\begin{tabular}[c]{@{}c@{}}Succ. rate\\ (\%)\end{tabular} & 100.00	& 99.93 & 100.00 &	100.00 & 100.00 &	99.86 &	99.93 \\ \hline
\begin{tabular}[c]{@{}c@{}}RMSE\\ position(cm)\end{tabular} & 0.17 & 0.12 & 0.18 & 0.13 & 0.09 & 0.29 & 0.14\\ \hline
\begin{tabular}[c]{@{}c@{}}RMSE\\ force(N)\end{tabular} & 0.01 & 0.01 & 0.01 & 0.01 & 0.01 & 0.02 & 0.02 \\ \hline
\end{tabular}%
}
\caption{Simulation Result : Single-contact case}
\label{tab:single_result}
\vspace{-5mm}
\end{table}

\subsection{Single contact case}
\label{subsec:simulation_single}
As shown in Table \ref{tab:overview}, on average, the success rate is 99.96\% with 0.16$\mathrm{cm}$, 0.01$\mathrm{N}$ RMSE. The algorithm run-time is 0.45$\mathrm{ms}$ and 18.25 steps are required to converge. Therefore, the algorithm estimates the contact in 8.21$\mathrm{ms}$ on average after the force is applied. We would like to underline that, the run-time of MCP-EP is at least $\geq3.5$ times faster compared to the existing methods \cite{popov2021real, zwiener2019armcl, manuelli2016localizing, popov2019real}. More detailed data can be found in Table \ref{tab:single_result}.

\begin{figure}[tb!]
    \centering
    \subfigure[]
	{\includegraphics[width=0.215\textwidth]{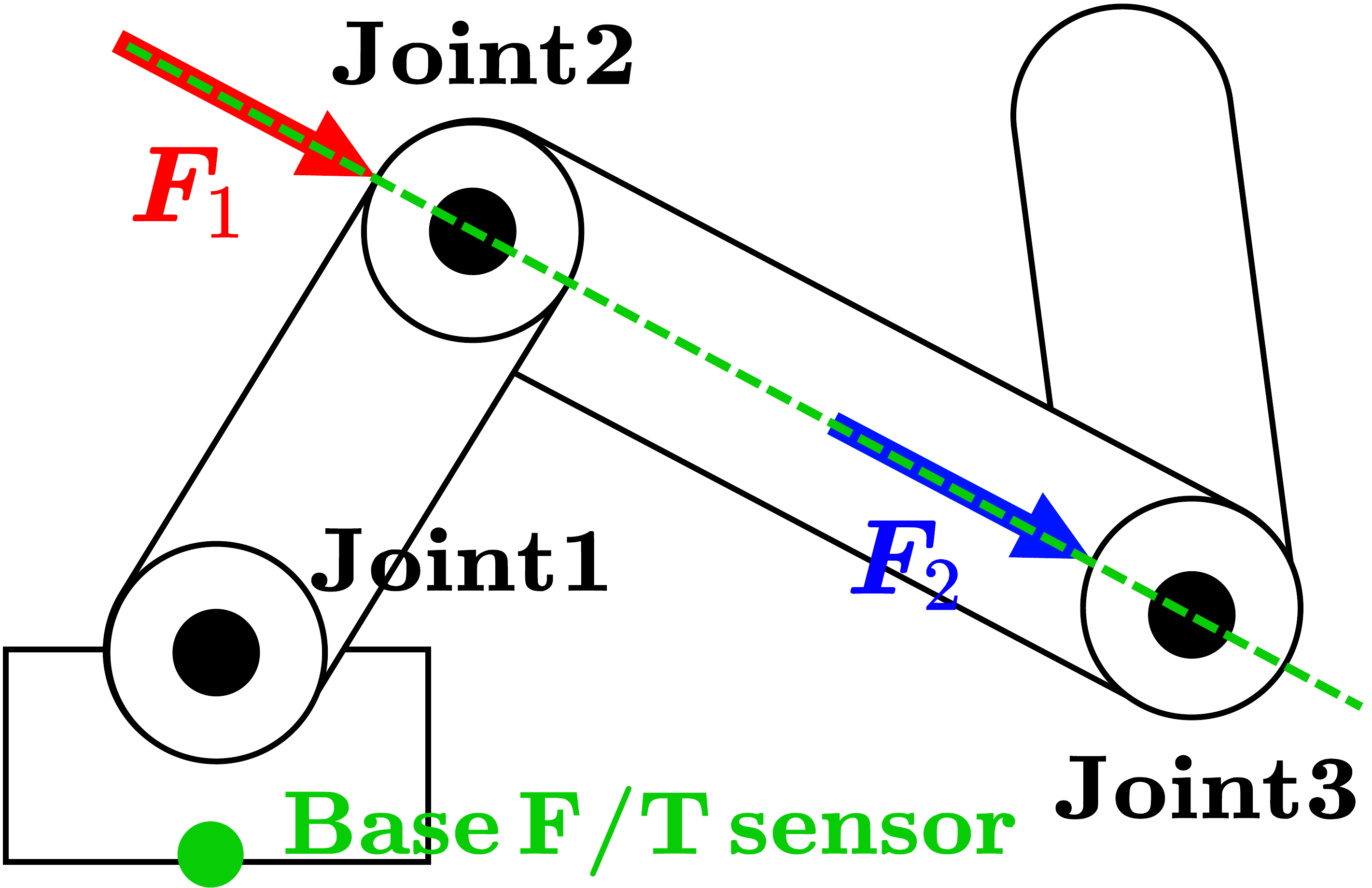}}
	\subfigure[]
	{\includegraphics[width=0.215\textwidth]{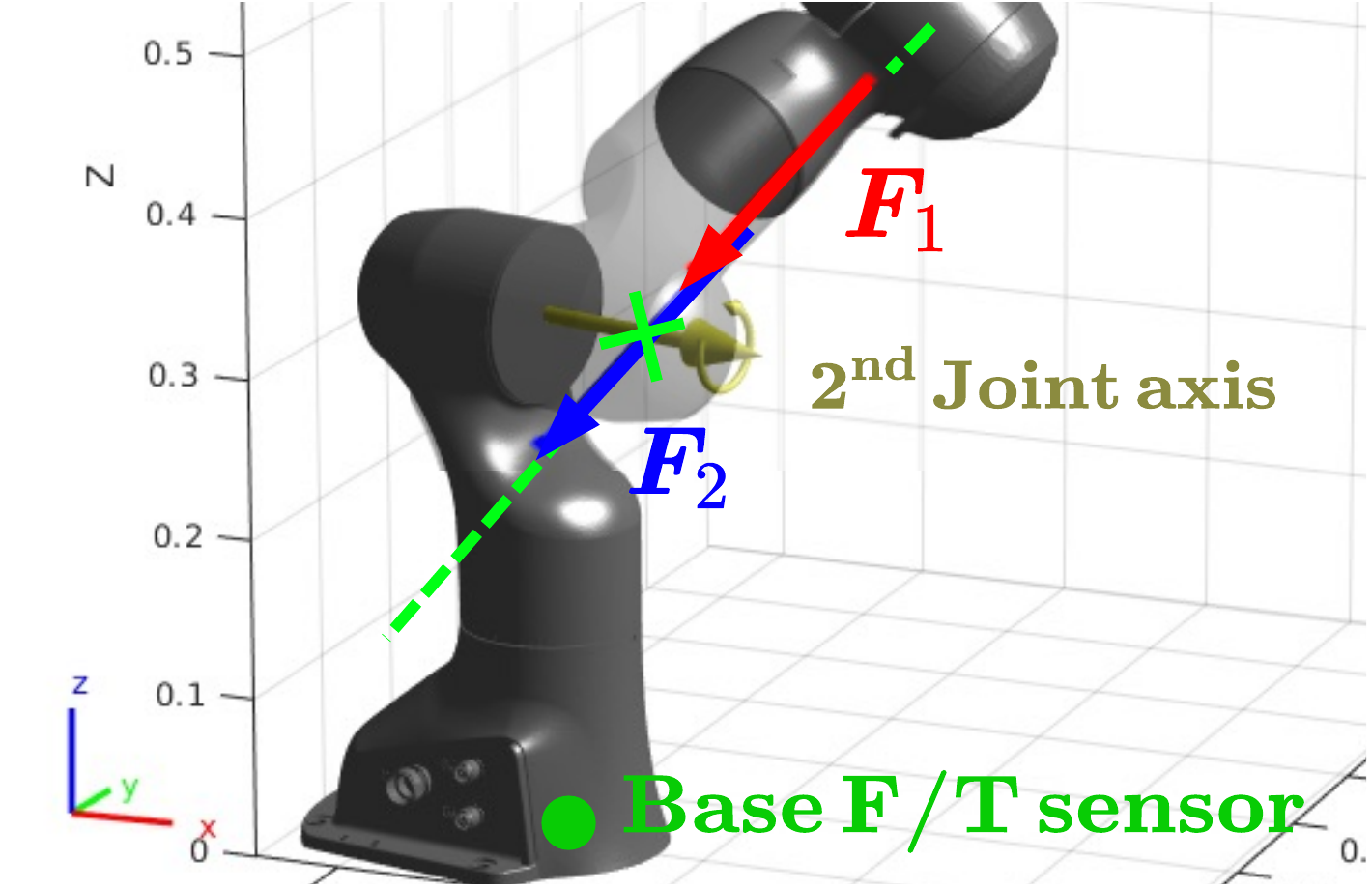}}
    \caption{
    (a) A schematic diagram that shows a typical singular case under single-contact. $\boldsymbol{F}_{1}$ and $\boldsymbol{F}_{2}$ cannot be distinguished because they produce the same sensor measurements.
    (b) A singular case for a 7-DOF robot.
    } 
    \label{fig:single_singular}
    \vspace{-6mm}
\end{figure}

Out of 10000 trials, contact point localization failed only 4 times, all due to the singularities. Although the use of the base F/T sensor largely mitigates the singularity, there are still some cases, in which multiple point and force pairs can perfectly explain the given sensor measurements. For example, $\boldsymbol{F}_1$ and $\boldsymbol{F}_2$ in Fig. \ref{fig:single_singular} produce the same sensor measurements, and as a consequence, the MCP-EP algorithm may fail. Nevertheless, apart from singularities, which we think are rare cases, single contact can be identified accurately. We remark that all RMSE position errors in $\mathrm{cm}$-scale in Table \ref{tab:single_result} corresponds to less than 1 face error (roughly speaking, 1 face error corresponds to 0.45$\mathrm{cm}$ error).

\begin{table}[thb!]
\resizebox{\columnwidth}{4.1cm}{%
\rotatebox[origin=c]{90}{{\normalsize \textbf{1st contact link}}}
\renewcommand{\arraystretch}{1.2}
\begin{tabular}{cclllllll}
\multicolumn{9}{c}{{\normalsize \textbf{2nd contact link}}} \\ \hline 
 & \textbf{} & \multicolumn{1}{c}{\textbf{link1}} & \multicolumn{1}{c}{\textbf{link2}} & \multicolumn{1}{c}{\textbf{link3}} & \multicolumn{1}{c}{\textbf{link4}} & \multicolumn{1}{c}{\textbf{link5}} & \multicolumn{1}{c}{\textbf{link6}} & \multicolumn{1}{c}{\textbf{link7}} \\ \hhline{=========}
\multirow{3}{*}{\textbf{link1}} & Succ. rate(\%) & \textbf{X} & 97.84 & 95.2 & 98.91 & 100.00 & 97.96 & 98.71 \\ \cline{2-9} 
 &RMSE(cm) & \textbf{X} & 0.99 & 1.46 & 0.62 & 0.44 & 0.92 & 0.38 \\ \cline{2-9} 
 & RMSE(N) & \textbf{X} & 3.09 & 2.22 & 1.33 & 0.54 & 0.68 & 0.27 \\ \hhline{=========}
\multirow{3}{*}{\textbf{link2}} &  Succ. rate(\%)   & 95.91 & \textbf{X} & 95.8 & 98.73 & 96.92 & 97.08 & 97.73 \\ \cline{2-9} 
 &  RMSE(cm)   & 1.50 & \textbf{X} & 1.70 & 1.14 & 1.18 & 0.88 & 0.88 \\ \cline{2-9} 
 &  RMSE(N)   & 3.62 & \textbf{X} & 4.00 & 2.52 & 1.36 & 0.88 & 0.68 \\ \hhline{=========}
\multirow{3}{*}{\textbf{link3}} &  Succ. rate(\%)   & 97.03 & 95.42 & \textbf{X} & 94.07 & 98.69 & 98.04 & 98.21 \\ \cline{2-9} 
 &  RMSE(cm)   & 1.37 & 1.33 & \textbf{X} & 1.91 & 1.03 & 0.97 & 0.80 \\ \cline{2-9} 
 &  RMSE(N)   & 2.68 & 3.59 & \textbf{X} & 4.18 & 1.63 & 1.37 & 0.91 \\ \hhline{=========}
\multirow{3}{*}{\textbf{link4}} &  Succ. rate(\%)   & 97.17 & 96.75 & 96.04 & \textbf{X} & 93.39 & 95.02 & 95.71 \\ \cline{2-9} 
 &  RMSE(cm)   & 0.89 & 0.85 & 1.41 & \textbf{X} & 1.82 & 1.71 & 1.32 \\ \cline{2-9} 
 &  RMSE(N)   & 1.33 & 1.69 & 3.85 & \textbf{X} & 3.83 & 2.82 & 1.79 \\ \hhline{=========}
\multirow{3}{*}{\textbf{link5}} &  Succ. rate(\%)   & 97.49 & 97.7 & 96.81 & 91.91 & \textbf{X} & 96.36 & 97.17 \\ \cline{2-9} 
 &  RMSE(cm)   & 1.01 & 1.06 & 1.18 & 1.49 & \textbf{X} & 1.42 & 0.89 \\ \cline{2-9}
 &  RMSE(N)   & 0.76 & 1.05 & 2.32 & 3.77 & \textbf{X} & 3.44 & 1.91 \\ \hhline{=========}
\multirow{3}{*}{\textbf{link6}} &  Succ. rate(\%)   & 96.85 & 94.74 & 96.54 & 92.00 & 91.53 & \textbf{X} & 94.05 \\ \cline{2-9} 
 &  RMSE(cm)   & 0.77 & 1.01 & 0.84 & 1.92 & 1.67 & \textbf{X} & 1.20 \\ \cline{2-9} 
 &  RMSE(N)   & 0.64 & 0.97 & 1.07 & 2.40 & 3.55 & \textbf{X} & 4.05 \\ \hhline{=========}
\multirow{3}{*}{\textbf{link7}} &  Succ. rate(\%)   & 99.55 & 98.82 & 99.61 & 98.71 & 97.61 & 97.91 & \textbf{X} \\ \cline{2-9} 
 &  RMSE(cm)   & 0.35 & 0.36 & 0.42 & 0.64 & 1.00 & 0.88 & \textbf{X} \\ \cline{2-9} 
 &  RMSE(N)   & 0.27 & 0.35 & 0.54 & 0.90 & 1.66 & 3.54 & \textbf{X} \\ \hline
\end{tabular}%
}
\caption{Simulation Result : Dual-contact case}
\label{tab:dual_result}
\vspace{-12mm}
\end{table}

\subsection{Dual contact case}
\label{subsec:simulation_dual}

To simulate the dual-contact scenario, forces are sequentially applied to the robot, with   ~0.1$\mathrm{s}$ time interval. The algorithm run-time is, on average, 1.66$\mathrm{ms}$ and 11 steps are required to converge (this value indicates the number of steps after the second particle set is added). Therefore, the MCP-EP estimates the second contact in 18.26$\mathrm{ms}$ on average after the second force is applied.

\begin{figure}[tb!]
    \centering
    \subfigure[]
	{\includegraphics[width=0.235\textwidth]{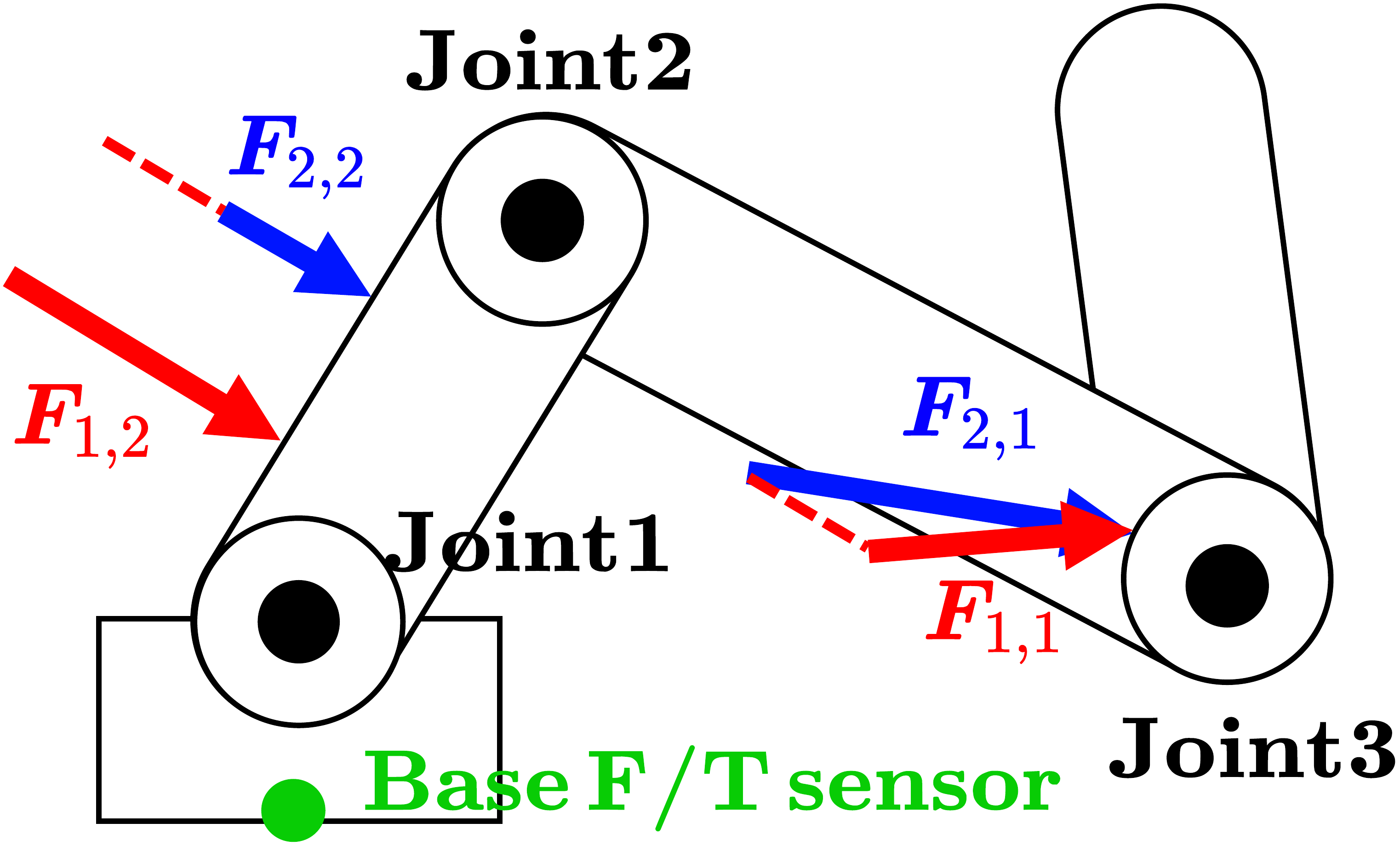}}
	\subfigure[]
	{\includegraphics[width=0.235\textwidth]{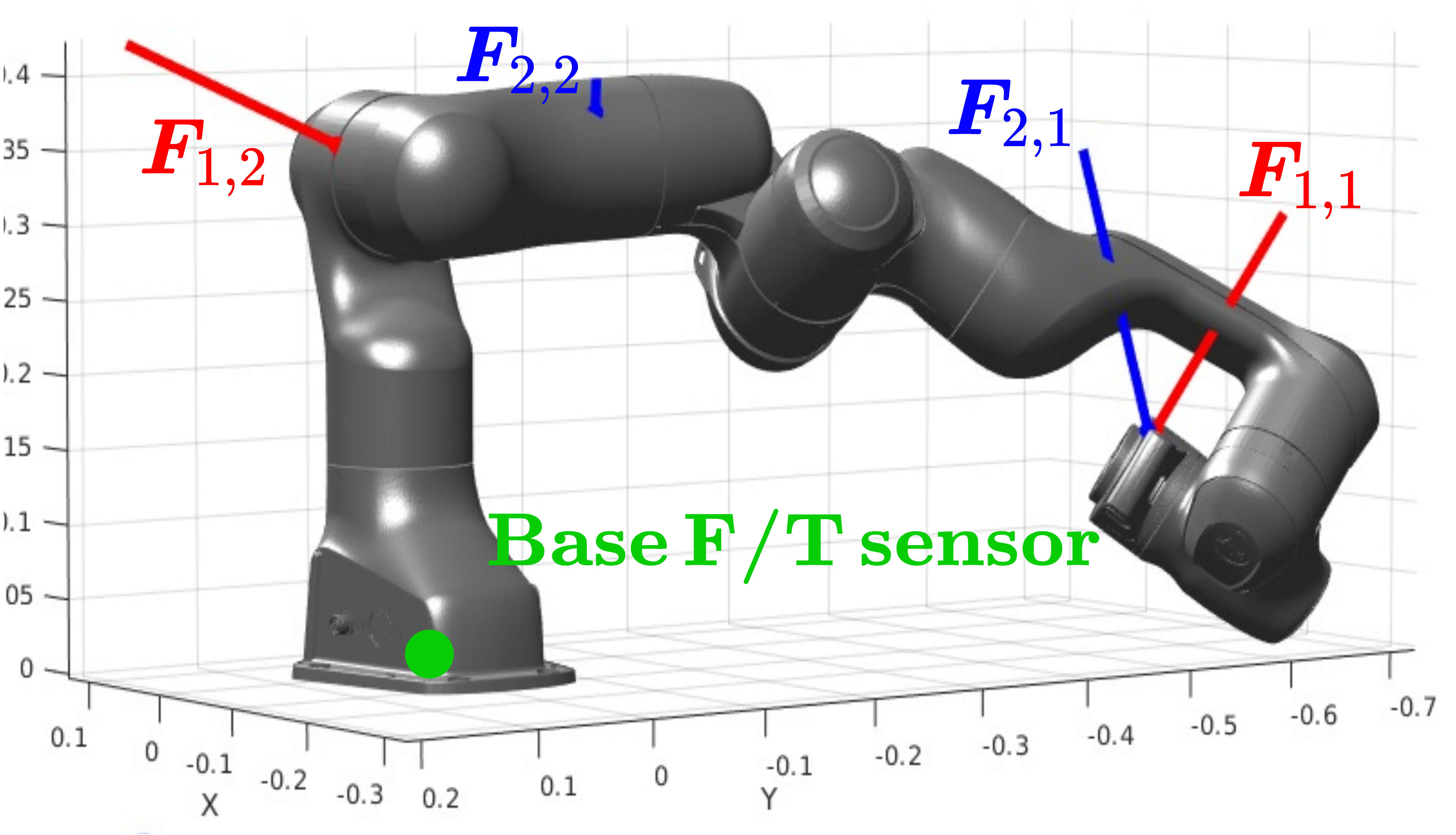}}
    \caption{
    (a) A schematic diagram that shows a singular case under dual-contact. Red and blue arrows will produce the very similar sensor measurements. 
    (b) A singular case for a 7-DOF robot.
    }
    \label{fig:dual_singular}
    \vspace{-6mm}
\end{figure}

As shown in Table \ref{tab:overview}, the overall success rate is 96.70\%. In our analysis, most of the failure cases fall into one of the following two categories. For the first category, one contact is successfully estimated, but the other contact falls into the singularity presented in Fig. \ref{fig:single_singular}. For the second category, the MCP-EP algorithm converges to a wrong estimate which still explains sensor measurements fairly well (hence local minima). In fact, since the first category is rare as discussed previously, most of failures are of the second category. Fig. \ref{fig:dual_singular} shows an example of the second category. Since the red and blue arrows may result in very similar sensor measurements (although they do not result in exactly the same values), MCP-EP algorithm may converge to a wrong one. 

From the above observation, the success rate of 96.70\% seems a fairly plausible value, as the dual-contact case has a higher chance of being singular. Moreover, in Table \ref{tab:dual_result}, we can find a tendency that, as the first and second contact links are farther, the success rate increases. This is intuitively acceptable, because, with rich sensory information between two contact points, the algorithm is less likely to fall into the singularity. We also remark that, when the first contact is made on the link 7, which would be one of the most common scenarios in practice, the success rate is 98.70\% on average.

\section{Conclusion}
\label{sec:conclusion}
In this work, we propose an algorithm called MCP-EP (Multi-Contact Particle Filter with Exploration Particles) to solve a multi-contact point localization and force identification problem. In the proposed algorithm, the use of the base F/T sensor enables to estimate multi contacts with little singularity. In validation, we presented estimation of three contacts for a 7-DOF robot (up to one contact per link). The proposed algorithm, in addition, improves the run-time using a proper mesh preprocessing, in which distance information for all face pairs are precomputed. The preprocessed data, however is valid only within a link, and consequently, particles are not updated across links through the motion model of the PF. This is overcome by modifying the PF algorithm with the concept of \textit{exploration particles} that are forced to investigate adjacent links. Simulation studies are presented to show the effectiveness of the proposed algorithm.

\bibliographystyle{IEEEtran}
\bibliography{IEEEabrv,[bib]ICRA2023_collision_detection}

\begin{thebibliography}{10}
\providecommand{\url}[1]{#1}
\csname url@rmstyle\endcsname
\providecommand{\newblock}{\relax}
\providecommand{\bibinfo}[2]{#2}
\providecommand\BIBentrySTDinterwordspacing{\spaceskip=0pt\relax}
\providecommand\BIBentryALTinterwordstretchfactor{4}
\providecommand\BIBentryALTinterwordspacing{\spaceskip=\fontdimen2\font plus
\BIBentryALTinterwordstretchfactor\fontdimen3\font minus
  \fontdimen4\font\relax}
\providecommand\BIBforeignlanguage[2]{{%
\expandafter\ifx\csname l@#1\endcsname\relax
\typeout{** WARNING: IEEEtran.bst: No hyphenation pattern has been}%
\typeout{** loaded for the language `#1'. Using the pattern for}%
\typeout{** the default language instead.}%
\else
\language=\csname l@#1\endcsname
\fi
#2}}

\bibitem{ebert2002safe}
D.~M. Ebert and D.~D. Henrich, ``Safe human-robot-cooperation: Image-based
  collision detection for industrial robots,'' in \emph{IEEE/RSJ international
  conference on intelligent robots and systems}, vol.~2, 2002, pp. 1826--1831.

\bibitem{kuhn2007fast}
S.~Kuhn and D.~Henrich, ``Fast vision-based minimum distance determination
  between known and unkown objects,'' in \emph{2007 IEEE/RSJ International
  Conference on Intelligent Robots and Systems}, pp. 2186--2191.

\bibitem{kim2023tamp}
K.~Kim and M.~J. Kim, ``A reachability tree-based algorithm for robot task and
  motion planning,'' in \emph{2023 IEEE International Conference on Robotics
  and Automation (ICRA)}.

\bibitem{kim2019passivity}
M.~J. Kim, W.~Lee, J.~Y. Choi, G.~Chung, K.-L. Han, I.~S. Choi, C.~Ott, and
  W.~K. Chung, ``A passivity-based nonlinear admittance control with
  application to powered upper-limb control under unknown environmental
  interactions,'' \emph{IEEE/ASME Transactions on Mechatronics}, vol.~24,
  no.~4, pp. 1473--1484, 2019.

\bibitem{kim2016passivity}
M.~J. Kim, W.~Lee, C.~Ott, and W.~K. Chung, ``A passivity-based admittance
  control design using feedback interconnections,'' in \emph{2016 IEEE/RSJ
  International Conference on Intelligent Robots and Systems (IROS)}, pp.
  801--807.

\bibitem{de2006collision}
A.~De~Luca, A.~Albu-Schaffer, S.~Haddadin, and G.~Hirzinger, ``Collision
  detection and safe reaction with the dlr-iii lightweight manipulator arm,''
  in \emph{2006 IEEE/RSJ International Conference on Intelligent Robots and
  Systems}, pp. 1623--1630.

\bibitem{kim2015design}
M.~J. Kim, Y.~J. Park, and W.~K. Chung, ``Design of a momentum-based
  disturbance observer for rigid and flexible joint robots,'' \emph{Intelligent
  Service Robotics}, vol.~8, pp. 57--65, 2015.

\bibitem{heo2019collision}
Y.~J. Heo, D.~Kim, W.~Lee, H.~Kim, J.~Park, and W.~K. Chung, ``Collision
  detection for industrial collaborative robots: A deep learning approach,''
  \emph{IEEE Robotics and Automation Letters}, vol.~4, no.~2, pp. 740--746,
  2019.

\bibitem{birjandi2020observer}
S.~A.~B. Birjandi, J.~K{\"u}hn, and S.~Haddadin, ``Observer-extended direct
  method for collision monitoring in robot manipulators using proprioception
  and imu sensing,'' \emph{IEEE Robotics and Automation Letters}, vol.~5,
  no.~2, pp. 954--961, 2020.

\bibitem{kim2021transferable}
D.~Kim, D.~Lim, and J.~Park, ``Transferable collision detection learning for
  collaborative manipulator using versatile modularized neural network,''
  \emph{IEEE Transactions on Robotics}, 2021.

\bibitem{park2021collision}
K.~M. Park, Y.~Park, S.~Yoon, and F.~C. Park, ``Collision detection for robot
  manipulators using unsupervised anomaly detection algorithms,''
  \emph{IEEE/ASME Transactions on Mechatronics}, 2021.

\bibitem{8695719}
J.~M. Gandarias, A.~J. García-Cerezo, and J.~M. Gómez-de Gabriel, ``Cnn-based
  methods for object recognition with high-resolution tactile sensors,''
  \emph{IEEE Sensors Journal}, vol.~19, no.~16, pp. 6872--6882, 2019.

\bibitem{8932392}
J.~Liang, J.~Wu, H.~Huang, W.~Xu, B.~Li, and F.~Xi, ``Soft sensitive skin for
  safety control of a nursing robot using proximity and tactile sensors,''
  \emph{IEEE Sensors Journal}, vol.~20, no.~7, pp. 3822--3830, 2020.

\bibitem{zwiener2018contact}
A.~Zwiener, C.~Geckeler, and A.~Zell, ``Contact point localization for
  articulated manipulators with proprioceptive sensors and machine learning,''
  in \emph{2018 IEEE International Conference on Robotics and Automation
  (ICRA)}, pp. 323--329.

\bibitem{8059840}
S.~Haddadin, A.~De~Luca, and A.~Albu-Schäffer, ``Robot collisions: A survey on
  detection, isolation, and identification,'' \emph{IEEE Transactions on
  Robotics}, vol.~33, no.~6, pp. 1292--1312, 2017.

\bibitem{buondonno2016combining}
G.~Buondonno and A.~De~Luca, ``Combining real and virtual sensors for measuring
  interaction forces and moments acting on a robot,'' in \emph{2016 IEEE/RSJ
  International Conference on Intelligent Robots and Systems (IROS)}, pp.
  794--800.

\bibitem{iskandar2021collision}
M.~Iskandar, O.~Eiberger, A.~Albu-Sch{\"a}ffer, A.~De~Luca, and A.~Dietrich,
  ``Collision detection, identification, and localization on the dlr sara robot
  with sensing redundancy,'' in \emph{2021 IEEE International Conference on
  Robotics and Automation (ICRA)}, pp. 3111--3117.

\bibitem{pang2021identifying}
T.~Pang, J.~Umenberger, and R.~Tedrake, ``Identifying external contacts from
  joint torque measurements on serial robotic arms and its limitations,'' in
  \emph{2021 IEEE International Conference on Robotics and Automation (ICRA)},
  pp. 6476--6482.

\bibitem{popov2021real}
D.~Popov, A.~Klimchik, and A.~Pashkevich, ``Real-time estimation of multiple
  potential contact locations and forces,'' \emph{IEEE Robotics and Automation
  Letters}, vol.~6, no.~4, pp. 7025--7032, 2021.

\bibitem{manuelli2016localizing}
L.~Manuelli and R.~Tedrake, ``Localizing external contact using proprioceptive
  sensors: The contact particle filter,'' in \emph{2016 IEEE/RSJ International
  Conference on Intelligent Robots and Systems (IROS)}, pp. 5062--5069.

\bibitem{zwiener2019armcl}
A.~Zwiener, R.~Hanten, C.~Schulz, and A.~Zell, ``Armcl: Arm contact point
  localization via monte carlo localization,'' in \emph{2019 IEEE/RSJ
  International Conference on Intelligent Robots and Systems (IROS)}, pp.
  7105--7111.

\bibitem{bimbo2019collision}
J.~Bimbo, C.~Pacchierotti, N.~G. Tsagarakis, and D.~Prattichizzo, ``Collision
  detection and isolation on a robot using joint torque sensing,'' in
  \emph{2019 IEEE/RSJ International Conference on Intelligent Robots and
  Systems (IROS)}, pp. 7604--7609.

\bibitem{popov2019real}
D.~Popov and A.~Klimchik, ``Real-time external contact force estimation and
  localization for collaborative robot,'' in \emph{2019 IEEE International
  Conference on Mechatronics (ICM)}, vol.~1, pp. 646--651.

\bibitem{likar2014external}
N.~Likar and L.~{\v{Z}}lajpah, ``External joint torque-based estimation of
  contact information,'' \emph{International Journal of Advanced Robotic
  Systems}, vol.~11, no.~7, p. 107, 2014.

\bibitem{haddadin2008collision}
S.~Haddadin, A.~Albu-Schaffer, A.~De~Luca, and G.~Hirzinger, ``Collision
  detection and reaction: A contribution to safe physical human-robot
  interaction,'' in \emph{2008 IEEE/RSJ International Conference on Intelligent
  Robots and Systems}, pp. 3356--3363.

\bibitem{Hess:2010:BFE:1893021}
R.~Hess, \emph{Blender Foundations: The Essential Guide to Learning Blender
  2.6}.\hskip 1em plus 0.5em minus 0.4em\relax Focal Press, 2010.

\bibitem{yan2009isotropic}
D.-M. Yan, B.~L{\'e}vy, Y.~Liu, F.~Sun, and W.~Wang, ``Isotropic remeshing with
  fast and exact computation of restricted voronoi diagram,'' in \emph{Computer
  graphics forum}, vol.~28, no.~5.\hskip 1em plus 0.5em minus 0.4em\relax Wiley
  Online Library, 2009, pp. 1445--1454.

\bibitem{gptoolbox}
A.~Jacobson \emph{et~al.}, ``{gptoolbox}: Geometry processing toolbox,'' 2021,
  http://github.com/alecjacobson/gptoolbox.

\bibitem{6386109}
E.~Todorov, T.~Erez, and Y.~Tassa, ``Mujoco: A physics engine for model-based
  control,'' in \emph{2012 IEEE/RSJ International Conference on Intelligent
  Robots and Systems}, pp. 5026--5033.

\bibitem{pandala2019qpswift}
A.~G. Pandala, Y.~Ding, and H.-W. Park, ``qpswift: A real-time sparse quadratic
  program solver for robotic applications,'' \emph{IEEE Robotics and Automation
  Letters}, vol.~4, no.~4, pp. 3355--3362, 2019.

\end{thebibliography}

\end{document}